\documentclass[compsoc,conference,a4paper,10pt,times]{IEEEtran}
\IEEEoverridecommandlockouts
\usepackage{cite}
\usepackage{amsthm}

\usepackage{algorithm}
\usepackage[noend]{algpseudocode} 
\usepackage{microtype}        
\usepackage{url}

\algrenewcommand\algorithmiccomment[1]{\hfill$\triangleright$~#1}

\usepackage{amsmath,amssymb,amsfonts}

\usepackage{graphicx}
\usepackage{bmpsize}
\usepackage{textcomp}
\usepackage{xcolor}
\usepackage{color, soul}
\usepackage{booktabs}
\usepackage{multirow}
\usepackage[colorlinks=true,urlcolor=black]{hyperref}
\usepackage[caption=false,font=footnotesize]{subfig}
\def\BibTeX{{\rm B\kern-.05em{\sc i\kern-.025em b}\kern-.08em
    T\kern-.1667em\lower.7ex\hbox{E}\kern-.125emX}}

\begin{document}

\title{Defending the Edge: Representative-Attention Defense against Backdoor Attacks in Federated Learning \thanks{This work was supported by EPSRC through the EnnCore project [EP/T026995/1]. Chibueze Peace Obioma was funded by Petroleum Technology Development Fund (PTDF), Abuja Nigeria, grant number PTDF/ED/OSS/PHD/CO/2079/22.}}

\author{
\IEEEauthorblockN{Chibueze Peace Obioma}
\IEEEauthorblockA{
\textit{Dept. of Computer Science}\\
\textit{Faculty of Engineering}\\
\textit{University of Manchester}\\
Manchester, United Kingdom\\
chibueze.obioma@manchester.ac.uk
}
\and
\IEEEauthorblockN{Youcheng Sun}
\IEEEauthorblockA{
\textit{Dept. of Computer Science}\\
\textit{Faculty of Engineering}\\
\textit{University of Manchester \& MBZUAI}\\
Abu Dhabi, United Arab Emirates\\
youcheng.sun@mbzuai.ac.ae
}
\and
\IEEEauthorblockN{Mustafa A. Mustafa}
\IEEEauthorblockA{
\textit{Dept. of Computer Science}\\
\textit{Faculty of Engineering}\\
\textit{University of Manchester \& KU Leuven}\\
Manchester, United Kingdom\\
mustafa.mustafa@manchester.ac.uk
}
}

\maketitle

\begin{abstract}
Federated learning (FL) remains highly vulnerable to adaptive backdoor attacks that preserve stealth by closely imitating benign update statistics. Existing defenses predominantly rely on anomaly detection in parameter or gradient space, overlooking behavioral constraints that backdoor attacks must satisfy to ensure reliable trigger activation. These anomaly-centric methods fail against adaptive attacks that normalize update magnitudes and mimic benign statistical patterns while preserving backdoor functionality, creating a fundamental detection gap. To address this limitation, this paper introduces FeRA (\underline{Fe}derated \underline{R}epresentative-\underline{A}ttention) -- a novel attention-driven defense that shifts the detection paradigm from anomaly-centric to consistency-centric analysis. FeRA exploits the intrinsic need for backdoor persistence across training rounds, identifying malicious clients through suppressed representation-space variance, an orthogonal property to traditional magnitude-based statistics. The framework conducts multi-dimensional behavioral analysis combining spectral and spatial attention, directional alignment, mutual similarity, and norm inflation across two complementary detection mechanisms: consistency analysis and norm-inflation detection. Through this mechanism, FeRA isolates malicious clients that exhibit low-variance consistency or magnitude amplification. Extensive evaluation across six datasets, nine attacks, and three model architectures under both independent and identically distributed (IID) and non-IID settings confirm FeRA achieves superior backdoor mitigation. Under different non-IID settings, FeRA achieves the lowest average backdoor accuracy (about 1.67\%) while maintaining high clean accuracy compared to other state-of-the-art defenses. The code is available at: \textit{\url{https://github.com/Peatech/FeRA_defense.git}}

\end{abstract}

\begin{IEEEkeywords}
Federated learning, backdoor attacks, Byzantine robustness, representation learning, attention mechanisms
\end{IEEEkeywords}

\section{Introduction}

Federated learning (FL) enables collaborative model training across distributed clients without centralizing sensitive data~\cite{mcmahan2017communication}. This architecture has found adoption in privacy-critical domains including healthcare diagnostics \cite{nguyen2022federated, silva2019federated}, financial fraud detection \cite{li2021survey}, and mobile keyboard prediction~\cite{kairouz2021advances}. However, the server's inability to inspect local data or verify computation integrity creates opportunities for Byzantine adversaries to inject malicious updates that compromise the global model.

Backdoor attacks present a distinct threat among Byzantine attacks. Unlike untargeted poisoning that degrades overall performance, backdoor attacks maintain high accuracy on benign inputs while embedding hidden behaviors triggered by specific patterns~\cite{bagdasaryan2020backdoor,wang2020attack}. An attacker selects a trigger pattern (e.g., a pixel patch) and target label, then poisons local training to associate the trigger with misclassification. The backdoored model satisfies dual objectives: high clean accuracy and reliable trigger activation. This stealth property enables persistence across training rounds while evading performance-based detection.

Most existing defenses operate from a premise that malicious behavior manifests as statistical anomalies \cite{cao2020fltrust, blanchard2017machine}. Byzantine clients are expected to exhibit deviations from the benign majority through gradient magnitude outliers, divergent parameter directions, or aberrant update patterns, leading detection methods to flag clients whose statistical signatures fall outside expected bounds. While some attacks~\cite{bagdasaryan2020backdoor, xie2019dba} do manifest as statistical outliers, this anomaly-centric perspective captures one dimension of the threat surface but overlooks a fundamental constraint: an attacker embedding a backdoor trigger must ensure reliable misclassification across hundreds of training rounds, despite model parameter drift, aggregation with benign updates, and stochastic client sampling. This reliability demands behavioral consistency. This consistency requirement creates detection surfaces beyond traditional anomaly metrics. 

The backdoor threat landscape exhibits significant diversity in how attacks manifest this consistency constraint. Some attacks~\cite{gu2019badnets, wang2020attack} prioritize reliability over stealth, accepting higher consistency signatures to ensure trigger activation. Others~\cite{xie2019dba, yang2024distributed} distribute the backdoor across colluding clients, fragmenting the consistency pattern but creating coordinated behavioral signatures. Adaptive attacks~\cite{fang2020local, zhang2023a3fl, wen2022thinking} calibrate poisoning intensity across rounds, trading reliability for reduced detectability, while scaling attacks~\cite{bagdasaryan2020backdoor} inject high-magnitude updates to dominate aggregation. This diversity reveals that different attack strategies ensure consistency through distinct mechanisms, each creating characteristic signatures across orthogonal dimensions. We empirically observe this pattern across multiple attack types (see Section~\ref{sec:method}), and this observation motivates taking a new detection approach targeting consistency patterns rather than anomalies alone. 

We propose FeRA (\underline{Fe}derated \underline{R}epresentative-\underline{A}ttention) -- a novel defense that identifies backdoor clients through multi-dimensional behavioral analysis. The term ``attention'' refers to weighted variance measures in representation space, specifically, how variance is distributed across spectral and spatial dimensions, rather than trainable attention mechanisms. We adapt terminology from visual attention models that identify salient features through variance concentration~\cite{itti2002model} to detect malicious variance suppression patterns that characterize backdoor attacks. FeRA computes six metrics capturing representation-space variance (spectral and spatial attention scores), parameter-space alignment (directional attention), coordination patterns (mutual similarity), and magnitude manipulation (spectral ratios). Detection operates through two complementary mechanisms addressing distinct backdoor characteristics. The consistency filter flags clients exhibiting simultaneously low variance representations, low directional attention, and high mutual similarity, capturing attacks that prioritize reliability through suppressed variance and coordinated patterns. The norm-inflation filter flags clients with inflated spectral norms, exposing scaling attacks that seek to dominate the aggregation process through oversized parameter updates. This design reflects a fundamental insight: backdoor attacks cannot simultaneously maintain effectiveness while evading variance-based detection and preserving benign-scale magnitudes.

\textbf{Contributions.} This work makes the following contributions to advancing backdoor defense in FL.
\begin{enumerate}
\item We identify the consistency pattern observed empirically across multiple backdoor attacks, demonstrating that reliability requirements create behavioral constraints beyond traditional anomaly metrics.

\item We introduce FeRA, a multi-dimensional defense framework that exploits these complementary constraints through two detection mechanisms: consistency analysis targeting suppressed representation variance and coordinated behavioral patterns, and norm-inflation detection identifying magnitude-amplified updates.

\item Through extensive evaluation across diverse attack types and realistic federated scenarios, we show FeRA achieves low backdoor accuracy (1.67\% mean value) while maintaining high clean accuracy (within 1.5\% of non-adversarial baselines), outperforming existing defenses.
\end{enumerate}

\section{Background and Related Work}
\label{sec:background}

This section provides essential background on FL security challenges. We survey backdoor attack methodologies, critically analyze existing defense mechanisms (particularly representation-based approaches), and identify fundamental vulnerabilities that motivate our defense design.

\subsection{Federated Learning and Security Challenges}

FL enables collaborative model training across distributed clients without centralizing sensitive data~\cite{mcmahan2017communication}. Given $K$ clients with local datasets $\{D_k\}_{k=1}^K$, FL seeks to minimize the global objective:
\begin{equation}
\min_{\theta} F(\theta) = \sum_{k=1}^{K} \frac{|D_k|}{|D|} F_k(\theta)
\end{equation}
\begin{equation}
F_k(\theta) = \frac{1}{|D_k|} \sum_{(x,y) \in D_k} \ell(\theta; x, y)
\end{equation}
where $\theta$ denotes model parameters, $\ell(\cdot)$ is the loss function, and $|D| = \sum_k |D_k|$. The standard FedAvg protocol~\cite{mcmahan2017communication} iteratively updates the global model through local training and aggregation: $\theta^{(t+1)} = \theta^{(t)} + \eta \sum_{k \in S_t} \frac{|D_k|}{|D|} \nabla F_k(\theta^{(t)})$, where $S_t$ denotes clients sampled at round $t$ and $\eta$ is the learning rate.

While FL preserves data privacy by design, its decentralized architecture introduces critical security vulnerabilities. The server cannot verify the integrity of client updates, creating opportunities for Byzantine attacks~\cite{blanchard2017machine} where malicious clients submit arbitrary model updates. Edge deployments exacerbate these risks through relaxed security policies, physical device access enabling hardware compromise, and intermittent participation patterns providing cover for strategic attack injection~\cite{kairouz2021advances}. Among Byzantine threats, backdoor attacks represent the most insidious challenge: unlike untargeted poisoning that degrades overall accuracy (easily detected), backdoors maintain high clean accuracy while triggering specific misclassifications on attacker-chosen inputs~\cite{bagdasaryan2020backdoor, wang2020attack}.

\subsection{Backdoor Attacks in Federated Learning}

Backdoor attacks embed hidden malicious behaviors into the global model through poisoned local updates. Formally, an attacker selects a trigger pattern $\tau$ (e.g., pixel patch, semantic perturbation) and target label $y_t$, then augments local training data: $D_k^- = D_k \cup \{(x + \tau, y_t) : (x, y) \in B_k\}$ where $B_k \subset D_k$ is the poisoned subset. The resulting backdoored model satisfies:
\begin{equation}
f_{\theta^*}(x) \approx y \quad \forall (x, y) \in D_{\text{clean}} \quad \text{(high clean accuracy)}
\end{equation}
\begin{equation}
f_{\theta^*}(x + \tau) \approx y_t \quad \forall x \quad \text{(high attack success)}
\end{equation}

Backdoor attacks in FL exhibit diverse sophistication levels. \textit{Model replacement attacks}~\cite{bagdasaryan2020backdoor} scale malicious updates by factor $\lambda = K/|M|$ where $K$ is total clients and $|M|$ is malicious count, aiming to dominate aggregation through magnitude manipulation. \textit{Distributed backdoor attacks (DBA)}~\cite{xie2019dba} decompose triggers across colluding clients, making individual updates appear benign while collectively embedding backdoors. \textit{Edge-case attacks}~\cite{wang2020attack} target rare inputs on distribution tails, exploiting natural class confusion patterns for highly stealthy backdoors. \textit{Neurotoxin}~\cite{zhang2022neurotoxin} achieves durability through momentum-guided parameter selection, slowly-changing parameters identified through gradient momentum analysis, ensuring backdoor persistence across hundreds of rounds. \textit{IBA}~\cite{nguyen2023iba} combines learnable trigger optimization with adversarial perturbations as triggers, making detection via visual inspection ineffective. Recent work explores stealthy attacks against personalized FL~\cite{lyu2024lurking}, data-free backdoor injection~\cite{li2024darkfed}, and distributed attacks on graph learning~\cite{yang2024distributed}. These adaptive attacks specifically design updates to satisfy defense constraints while preserving malicious functionality~\cite{wang2020attack}.

\subsection{Defenses Against Backdoor Attacks}

Defenses against backdoor attacks in FL operate across three temporal phases, each employing distinct detection strategies with characteristic vulnerabilities to adaptive attacks.

\textbf{Pre-aggregation defenses.} These methods filter malicious updates before aggregation by analyzing individual client contributions.  FoolsGold~\cite{fung2020foolsgold} exploit learning history to identify clients with similar gradient patterns, reducing weights for coordinated attackers. FLAME~\cite{nguyen2022flame} employs HDBSCAN clustering on client representations, excluding minority clusters as potentially malicious. DeepSight~\cite{rieger2022deepsight} applies spectral clustering on representation covariance matrices, and carries out deep model inspection to analyze internal representations for anomalous patterns. Trust bootstrapping methods like FLTrust~\cite{cao2020fltrust} maintain a small root dataset on the server, computing cosine similarity between client updates and trusted root updates to filter anomalous contributions. Other recent works that falls under this category include FedAvgCKA~\cite{walter2024exploiting}, frequency analysis method~\cite{fereidooni2024freqfed}, out-of-distribution data detection~\cite{li2024backdoorindicator}, federated detection~\cite{rieger2024crowdguard}, and certified robustness~\cite{xie2021crfl}.

\begin{figure}[t]
  \centering
  \includegraphics[width=0.7\columnwidth]{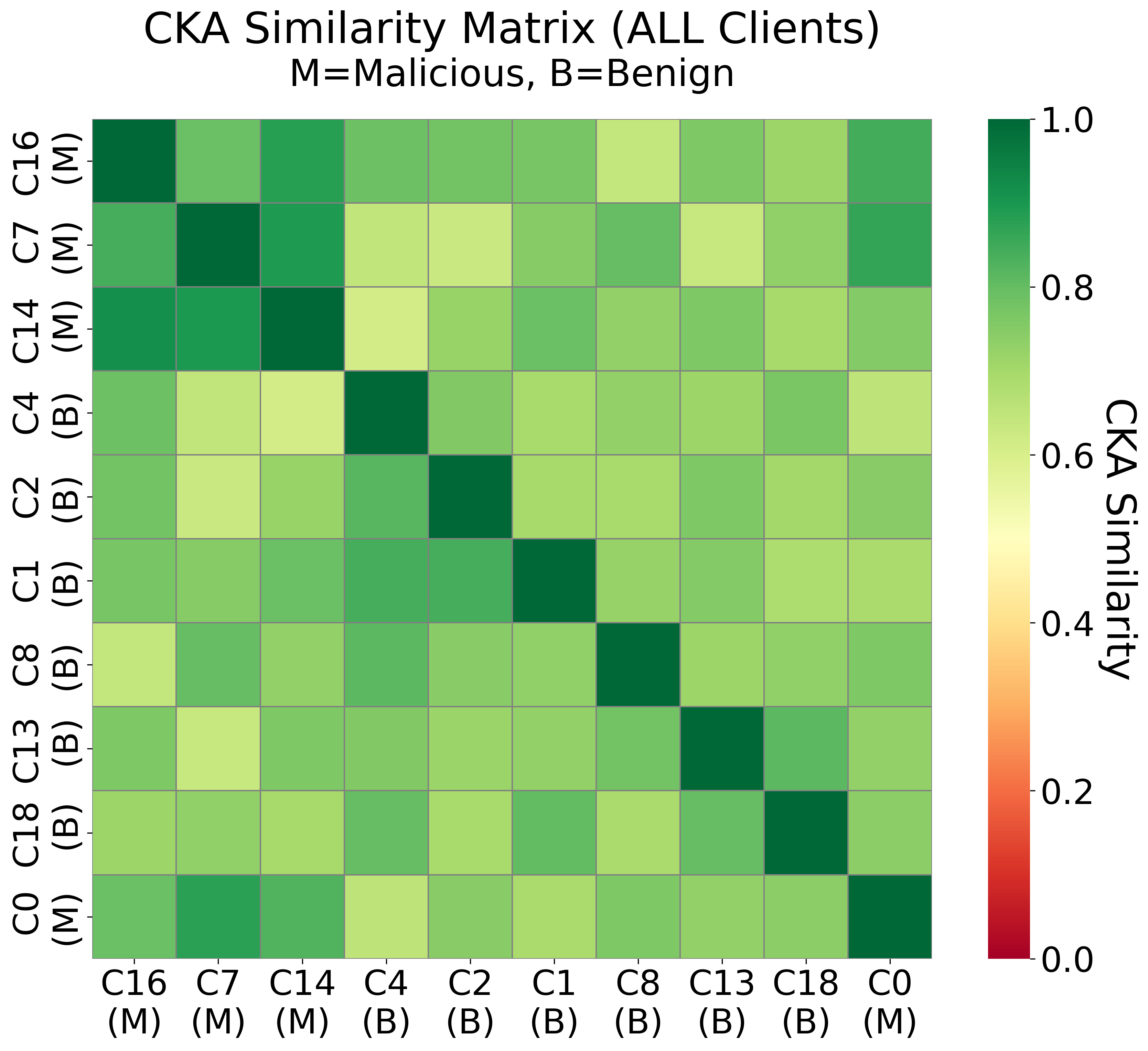}
  \caption{Centered kernel alignment (CKA) similarity matrix. Malicious clients increase their similarity to benign clients, reducing the separability required by metric-based detection approaches.}
  \label{fig:cka_attack}
\end{figure}

\begin{figure}[t]
  \centering
  \includegraphics[width=0.7\columnwidth]{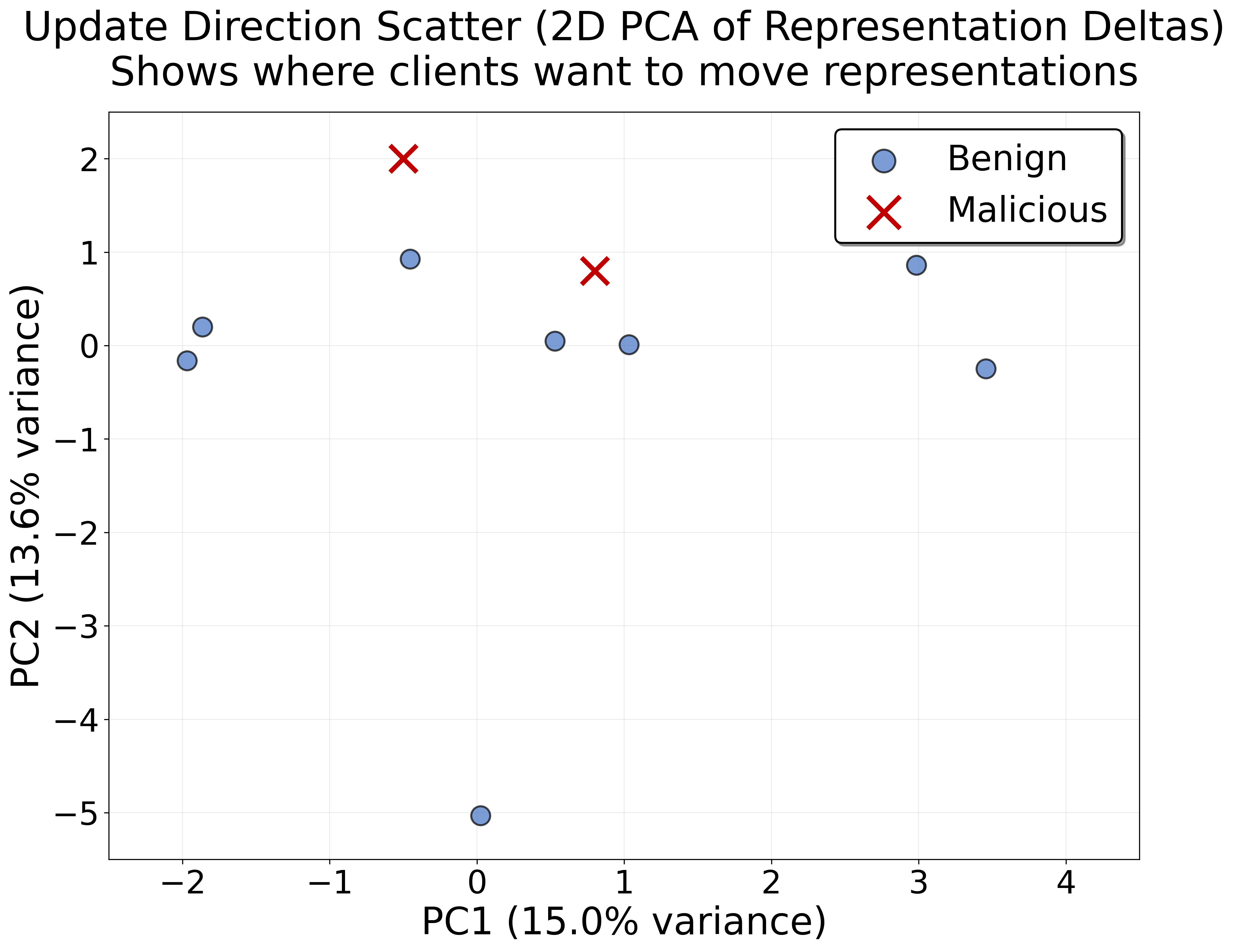}
  \caption{Two-dimensional principal component analysis (2D PCA) of representation deltas. Malicious updates overlap with benign representation shifts, reducing geometric separation and weakening clustering-based detection.}
  \label{fig:pca_scatter}
\end{figure}

However, these approaches face fundamental limitations. Defenses like FLTrust and FedAvgCKA~\cite{walter2024exploiting} that measure client similarity, are vulnerable to adaptive optimization. As illustrated in Fig.~\ref{fig:cka_attack}, attackers can craft updates that maximize centered kernel alignment (CKA) similarity or cosine distance while maintaining backdoor effectiveness~\cite{wang2020attack}, exploiting the single-metric nature of these defenses. Clustering methods fail when attackers position malicious representations within benign clusters through careful adversarial crafting. Fig.~\ref{fig:pca_scatter} demonstrates this coordinate adaptation vulnerability, where malicious clients (red) embed within benign clusters (blue), defeating spatial partitioning. Adaptive attackers systematically evade pre-aggregation filters by mimicking benign statistics~\cite{fang2020local}.

\textbf{In-aggregation defenses.} These methods apply Byzantine-robust aggregation rules to limit malicious influence during model combination. MKrum~\cite{blanchard2017machine} selects clients whose updates have minimal sum of squared distances to their $k$ nearest neighbors, assuming benign updates cluster together. Bulyan~\cite{mhamdi2018hidden} recursively applies Byzantine-robust selection rules like MKrum followed by coordinate-wise trimmed mean. Coordinate-wise median~\cite{yin2018byzantine} and geometric median aggregation~\cite{blanchard2017machine} provide provable robustness against untargeted attacks. Recent innovations include provable defense frameworks~\cite{zhang2023flip}, game-theoretic approaches~\cite{jia2023fedgame}, parameter disparity analysis for heterogeneous settings~\cite{huang2024parameter}, and direction alignment inspection~\cite{xu2025alignins}.

Despite theoretical guarantees against untargeted poisoning, these methods struggle with backdoor attacks. Sophisticated attackers normalize update magnitudes to appear benign while preserving backdoor functionality through careful parameter crafting. Adaptive attacks bypass robust aggregation through scaled-down updates~\cite{fang2020local}, motivating dynamic learning rate approaches~\cite{ozdayi2021defending}. Most of these defences operate in the parameter space, and fundamentally disregard semantic information encoded in learned representations, missing backdoor-specific patterns that manifest in feature space rather than parameter magnitude.

\textbf{Post-aggregation defenses.} These approaches remove backdoors after global model updates through model repair techniques. Trigger reverse-engineering~\cite{wang2019neural} attempts to identify and neutralize backdoor triggers through optimization. Neuron pruning and fine-tuning~\cite{liu2018finepruning} remove suspicious neurons followed by retraining on clean data. Neural attention distillation~\cite{liu2018finepruning} transfers knowledge from potentially backdoored models to clean student models. Federated unlearning~\cite{sun2019can} selectively removes the influence of suspected malicious contributions from the global model.

Post-aggregation methods require auxiliary clean data for trigger identification or fine-tuning, which may be unavailable in privacy-sensitive FL deployments. These approaches struggle with stealthy triggers that blend seamlessly with benign model behaviors, particularly for edge-case attacks~\cite{wang2020attack} that exploit natural class confusion. Once backdoors are embedded in the global model, complete removal becomes challenging without degrading main task performance.

\subsection{Critical Vulnerabilities in Existing Defenses}

We identify three fundamental vulnerabilities that underscore the design of FeRA. 

First, existing defenses rely on static detection assumptions, expecting malicious updates to exhibit consistent deviations from benign patterns under fixed metrics. Byzantine-robust aggregation rules such as MKrum~\cite{blanchard2017machine} and Bulyan~\cite{mhamdi2018hidden} operate through distance-based outlier detection and geometric median computation, achieving provable robustness against untargeted poisoning. However, sophisticated attackers normalize update magnitudes while preserving backdoor functionality through careful parameter crafting, violating the fixed-deviation assumption. Defenses like FLTrust~\cite{cao2020fltrust}, maintain a root dataset for trust bootstrapping, computing cosine similarity between client updates and trusted updates to filter anomalous contributions. Yet adaptive attackers exploit these single-metric approaches through coordinate-wise optimization, maximizing similarity scores while maintaining backdoor effectiveness \cite{bagdasaryan2020backdoor}. These defenses implicitly assume that malicious representations maintain fixed geometric relationships to benign ones, an assumption that collapses under adaptive manipulation.

Second, defenses that employ clustering exhibit structural fragility when attackers position malicious representations within benign clusters. FLAME \cite{nguyen2022flame} employs HDBSCAN clustering on cosine distances to identify and exclude minority clusters as potentially malicious, applying differential privacy-inspired noise injection to eliminate remaining backdoors. DeepSight \cite{rieger2022deepsight} uses spectral clustering on representation covariance matrices, introducing division differences (DDifs) and normalized energy updates (NEUPs) to characterize training data distributions. FoolsGold~\cite{fung2020foolsgold} exploits learning history to identify clients with similar gradient patterns, reducing weights for Sybil attackers through historical gradient similarity analysis. However, these approaches fail when attackers embed malicious representations in regions of high benign density through careful adversarial crafting. Clustering methods fundamentally depend on spatial separability, yet backdoor attacks can achieve reliable trigger activation while occupying the statistical interior of benign distributions.

Third, single-metric approaches create optimization opportunities where attackers can satisfy defense constraints while preserving malicious functionality. Prior work identifies that backdoors leave spectral signatures~\cite{tran2018spectral}, but our analysis reveals these signatures extend across multiple complementary analytical dimensions: geometric similarity through CKA, spectral structure via eigenvalue distributions, and directional alignment through cosine similarity. Current defenses examine these dimensions in isolation, enabling attackers to optimize against the specific metric under evaluation. This insight motivates FeRA's multi-dimensional detection framework, which constructs orthogonal detection surfaces that resist simultaneous normalization.

\section{Threat Model and Defense Objectives}
\label{sec:threat}

We formalize the adversarial setting for FL under backdoor attacks, establishing attacker capabilities, defender objectives, and system assumptions that guide our defense design.

\subsection{Attacker Model}

\textbf{Attacker goals.} The adversary aims to embed persistent backdoor functionality into the global model while maintaining stealth to evade detection. Formally, given trigger pattern $\tau$ and target label $y_t$, the attacker seeks a backdoored model $\theta^*$ satisfying:
\begin{equation}
\mathbb{E}_{(x,y) \sim D_{\text{clean}}} [\mathbb{I}[f_{\theta^*}(x) = y]] \geq 1 - \epsilon_{\text{clean}}
\end{equation}
\begin{equation}
\mathbb{E}_{x \sim D} [\mathbb{I}[f_{\theta^*}(x + \tau) = y_t]] \geq 1 - \epsilon_{\text{attack}}
\end{equation}
where $\epsilon_{\text{clean}}, \epsilon_{\text{attack}} \ll 1$ ensure the backdoor remains functional yet undetectable through accuracy degradation. The attacker prioritizes attack persistence (backdoor durability across training rounds) and detection evasion (avoiding identification by defense mechanisms).

\textbf{Attacker capabilities.} Following established threat models in Byzantine-robust FL~\cite{blanchard2017machine, bagdasaryan2020backdoor}, we assume the attacker controls a subset $\mathcal{M} \subset \{1, \ldots, K\}$ of malicious clients with $|\mathcal{M}| < K/2$, following recent work~\cite{zhang2022neurotoxin, wang2020attack}. Malicious clients have full local control of their local model. They craft adaptive updates to meet defense constraints while preserving backdoors~\cite{wang2020attack}.

\textbf{Attacker limitations.} We assume the attacker cannot compromise the server, cannot control the majority of clients, and cannot observe other clients' local data or models. However, we explicitly allow malicious clients to share and coordinate their updates with each other, as observed in distributed backdoor attacks~\cite{xie2019dba}. These assumptions are standard in FL security research~\cite{kairouz2021advances}.

\subsection{Defender Model and Assumptions}

\textbf{Defender capabilities.} The server analyzes client model updates using a small root dataset $D_{\text{root}}$ of size $|D_{\text{root}}| = 64$ drawn from the test set, with no access to local training data. Critically, $D_{\text{root}}$ need not match the federated training distribution $D = \bigcup_k D_k$; ablation studies (Section~\ref{sec:experiments}) demonstrate effectiveness even with out-of-distribution root data, enabling practical deployment when server-side data matching client distributions is unavailable. Representation extraction can be performed within a trusted execution environment to preserve client privacy, though our implementation does not require trusted execution environment (TEE) deployment. The server operates without prior knowledge of the number of malicious clients, backdoor triggers, or target labels. Clients train anonymously with untraceable individual actions.

\subsection{Defense Challenges}

Given this threat model, FeRA addresses three core challenges that existing defenses fail to overcome:

\textbf{C1: Establishing multi-dimensional detection barriers.} Most existing defenses operate on single behavioral dimensions, enabling attackers to optimize malicious updates against specific detection metrics while preserving backdoor functionality. The defense must construct detection surfaces across orthogonal behavioral dimensions where normalizing one metric necessarily degrades others. This creates optimization barriers with exponentially shrinking feasible regions as the number of constraints increases~\cite{huang2023multimetrics, yin2018byzantine}. The challenge lies in identifying behavioral dimensions that are mathematically complementary rather than merely correlated, ensuring that attacks cannot simultaneously satisfy all constraints without sacrificing effectiveness.

\textbf{C2: Achieving malicious client detection under non-IID heterogeneity.} Byzantine-robust aggregation requires complete identification of malicious clients to prevent backdoor injection. Under the minority Byzantine assumption ($|M| < 0.5N$), the defense must attain a high recall (eliminating false negatives) while maintaining acceptable precision under heterogeneous data distributions. The challenge is designing detection metrics where malicious clients consistently occupy identifiable regions of the behavioral space regardless of their adaptive strategies, enabling guaranteed detection without misclassifying benign heterogeneity as malicious behavior.

\textbf{C3: Preventing backdoor accumulation through aggregation design.} Binary accept-reject decisions create discontinuous detection boundaries where attacks positioned near decision thresholds achieve disproportionate influence when detection fails. Even with high detection accuracy, rare detection errors enable backdoor accumulation across rounds as undetected malicious updates compound through iterative aggregation. The defense must implement aggregation mechanisms that limit backdoor contribution even when detection is imperfect. This requires moving beyond binary classification to graduated trust mechanisms that modulate malicious influence through continuous weight functions, ensuring that near-threshold attacks cannot dominate aggregation through strategic positioning.

These challenges reflect fundamental limitations in current defense paradigms. The subsequent methodology section (Section~\ref{sec:method}) presents FeRA's technical approach, demonstrating how multi-dimensional consistency analysis addresses these challenges through complementary detection surfaces and inverted attentional filtering.

\section{FeRA Methodology}
\label{sec:method}
\subsection{Observations and Design Choices}
\label{sec:observation}

We investigate how backdoor injection manifests across multiple behavioral dimensions through controlled synthetic experiments. Following established backdoor methodologies \cite{bagdasaryan2020backdoor,wang2020attack,xie2019dba}, we constructed FL scenarios where Byzantine clients embed various trigger patterns into local training data while maintaining high clean accuracy to evade performance-based detection.

\begin{figure}[t]
\centering
\includegraphics[width=0.45\textwidth]{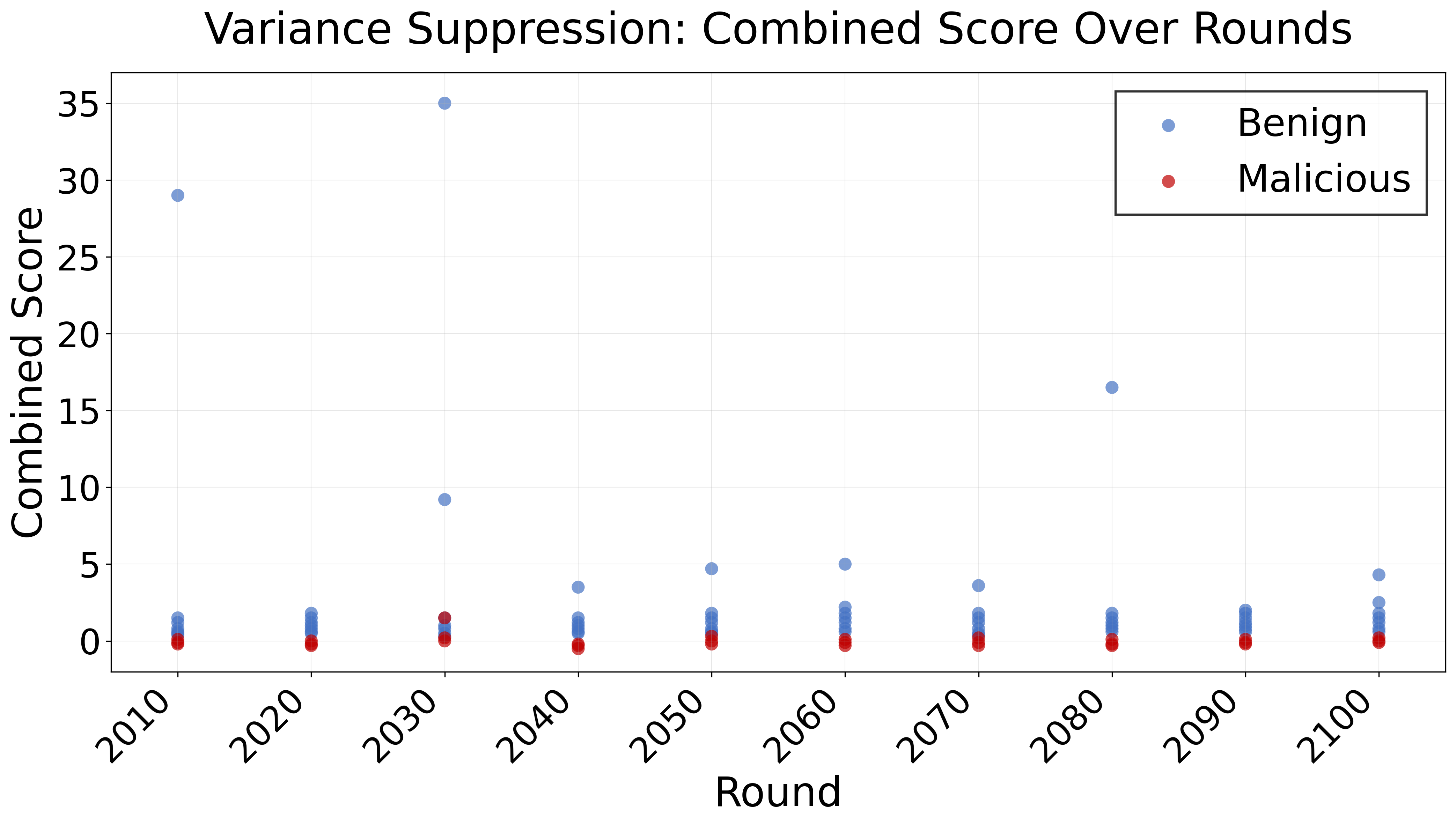}
\caption{Representation variance distribution across benign and backdoored clients across $n=100$ rounds.}
\label{fig:variance_suppression}
\end{figure}

\textbf{Observation 1: Variance suppression and consistency.} Across multiple attack scenarios, we observed that clients embedding backdoors exhibit systematically suppressed representation variance compared to benign clients. Fig.~\ref{fig:variance_suppression} illustrates this phenomenon across 100 rounds. 
Benign clients training on heterogeneous local data naturally produce diverse feature activations, reflected in high spectral variance ($\lambda_{\max}$ of centered covariance) and spatial dispersion (Frobenius norm). In contrast, backdoored clients demonstrate attentional collapse: representations concentrate in low-dimensional trigger-aligned subspaces.

%\textcolor{red}{[MM: Make sure Fig. 3 and 4 appear next to each other at the top of the page (one on each column) rather than both appearing one below the other on a single column.]}

\begin{figure}[t]
\centering
\includegraphics[width=0.4\textwidth]{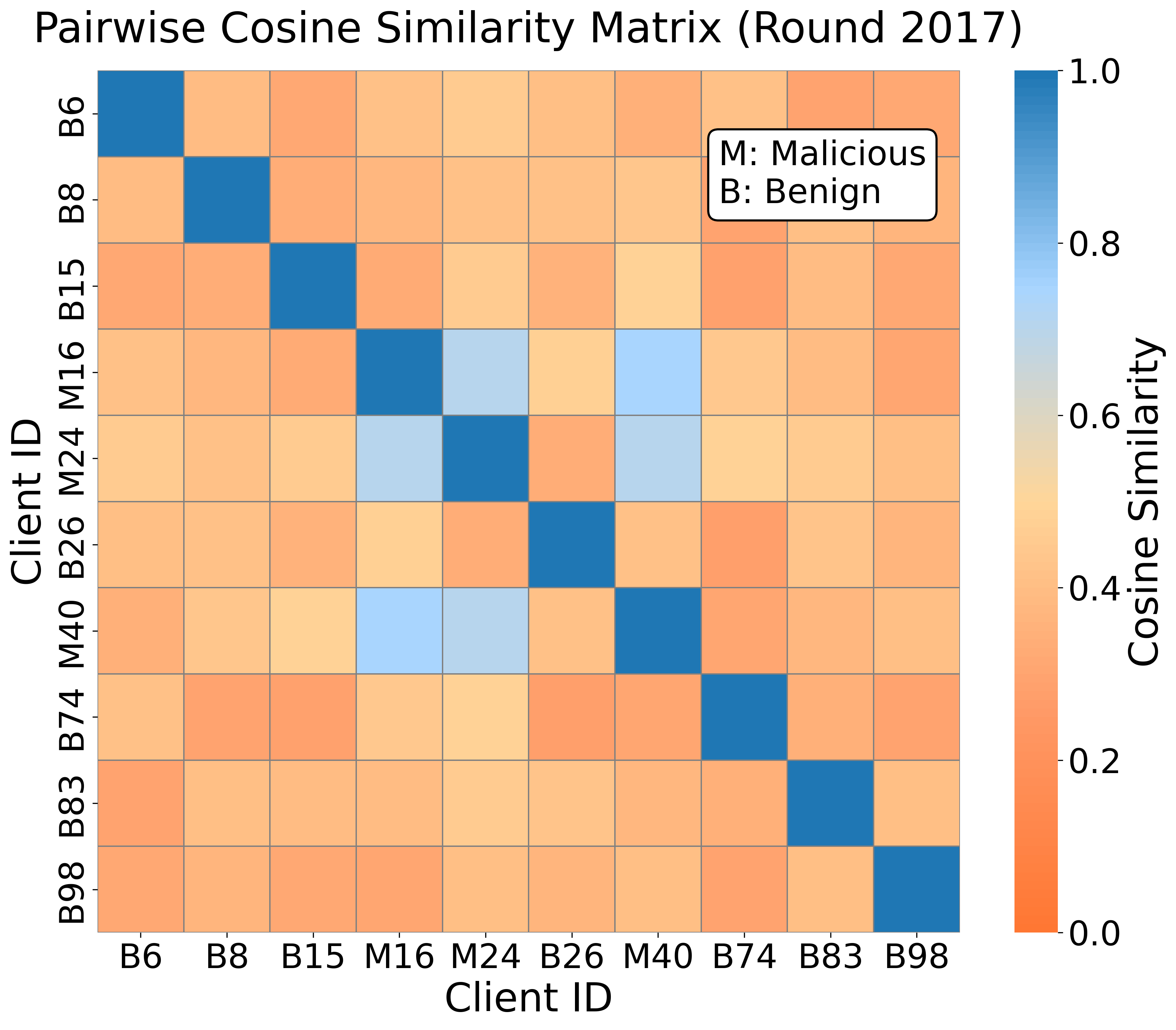}
\caption{Inter-client parameter similarity reveals coordination patterns. Heatmap shows pairwise cosine similarity for 10 clients (3 malicious, 7 benign) over 50 rounds. Malicious-malicious pairs (marked M) exhibit higher mean similarity compared to benign-benign pairs, indicating coordinated optimization toward shared backdoor objectives despite data heterogeneity.}
\label{fig:collusion_patterns}
\end{figure}

\begin{figure*}[ht]
\centering
\includegraphics[width=1\textwidth]{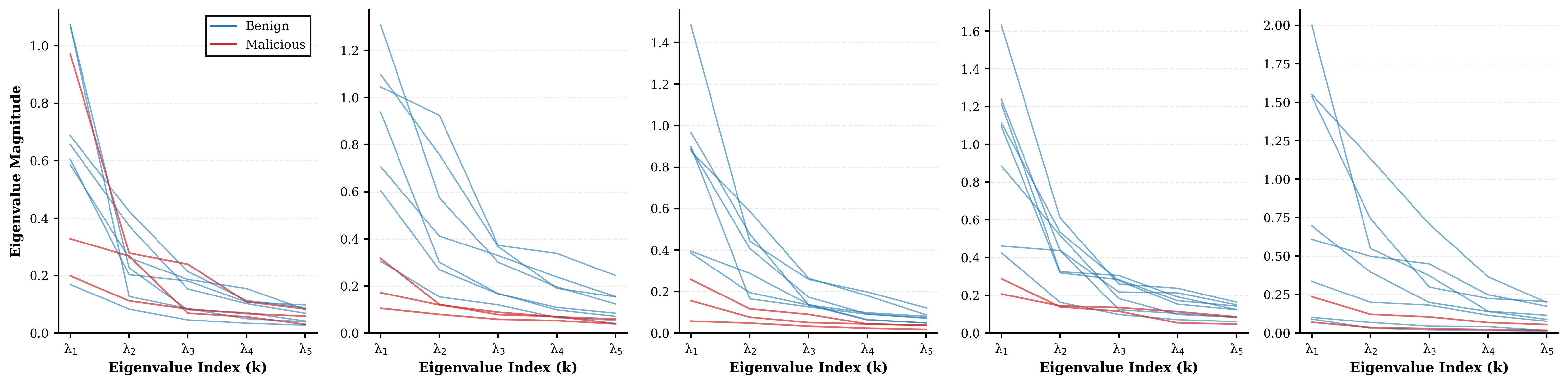}
\caption{Spectral variance suppression across top-$k$ eigenvalues. Malicious clients (red) exhibit consistently lower eigenvalues $\lambda_1$ through $\lambda_5$ compared to benign clients (blue) across several rounds under attack on CIFAR-10. 
}
\label{fig:topk_eigenvalues}
\end{figure*}

This variance suppression directly reflects backdoor effectiveness constraints. For reliable misclassification across hundreds of training rounds despite parameter drift and benign aggregation, trigger-associated representations must remain stable. Poisoned samples $(x + \tau, y_t)$ consistently map to restricted feature subspaces, suppressing variance along non-trigger dimensions. 

Critically, this suppression manifests across the dominant spectral subspace, not merely along the principal direction. Our analysis of the full eigenspectrum $\{\lambda_1, \lambda_2, \ldots, \lambda_k\}$ reveals that malicious clients exhibit systematically reduced eigenvalues across the top $k{=}5$ components (mean reduction of 68\% relative to benign clients, see Fig.~\ref{fig:topk_eigenvalues}). 
This multi-dimensional collapse indicates that backdoor objectives constrain learning across multiple orthogonal directions simultaneously, rather than through simple directional alignment. Benign clients face no such constraint, as heterogeneous data naturally induces diverse learning trajectories across all principal components.

Beyond variance suppression, we observe elevated inter-client behavioral coordination in scenarios where multiple malicious clients participate (e.g., distributed backdoor attacks \cite{xie2019dba}). Fig.~\ref{fig:collusion_patterns} demonstrates that colluding attackers, despite training on different local data partitions, exhibit systematically higher pairwise parameter similarity compared to benign client pairs. This coordination emerges because colluding attackers optimize identical backdoor objectives, mapping trigger pattern $\tau$ to target label $y_t$ regardless of local data differences \cite{fang2020local}. While benign clients' parameter trajectories diverge due to heterogeneous task distributions, malicious clients converge toward shared trigger-aligned representations. Critically, this pattern persists even when individual malicious updates are normalized to appear benign in magnitude or direction. The coordination signature manifests in the upper tail of mutual similarity metrics.

These observations motivate FeRA's \textit{Consistency Filter}, which identifies clients exhibiting simultaneously low combined scores (capturing both spectral and spatial variance suppression), low directional attention (deviation from global trajectory), and high mutual similarity (coordination patterns).

\begin{figure}[t]
\centering
\includegraphics[width=0.48\textwidth]{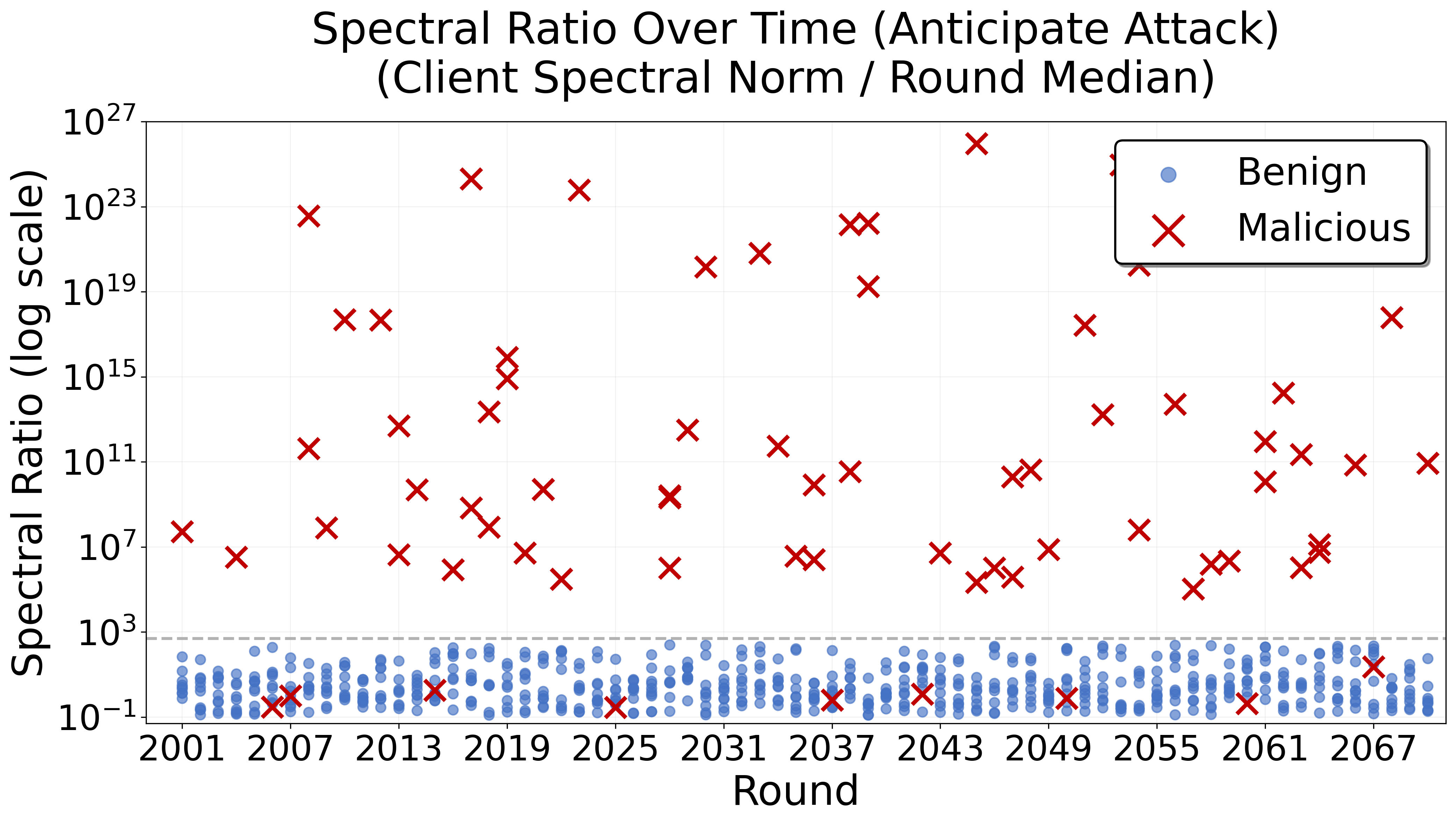}
\caption{Scaling attacks exhibit spectral ratio amplification. Time series shows spectral ratio for scaled attacks (red) vs. benign clients (blue).}
\label{fig:norm_inflation}
\end{figure}

\textbf{Observation 2: Magnitude amplification in scaling attacks.} Certain attack strategies dominate aggregation through explicit parameter norm inflation. Fig.~\ref{fig:norm_inflation} shows that scaling attacks \cite{bagdasaryan2020backdoor} scale malicious updates to ensure backdoor dominance despite benign majority.

While magnitude-agnostic metrics such as directional alignment and mutual similarity often fail to expose this class of manipulations, spectral decomposition provides a more transparent view of the update dynamics. By examining the dominant singular value of the largest network layer, we observed that benign client updates exhibit tightly bounded spectral behavior consistent with normal optimization dynamics, reflecting stable learning rates and gradient magnitudes, whereas maliciously perturbed updates consistently distort this spectral profile, producing measurable deviations in the leading singular mode of $\theta_{(i)}$ relative to $\theta_{(global)}$. This observation motivates FeRA's \emph{Norm-Inflation Filter}, which flags clients whose spectral ratios exceed benign bounds.

Our empirical findings were synthesized into a coherent detection framework as presented in the methodology section with details of the algorithmic implementation.

\subsection{FeRA Overview}
\label{method}
\begin{figure}[t]
\centering
\includegraphics[width=0.5\textwidth]{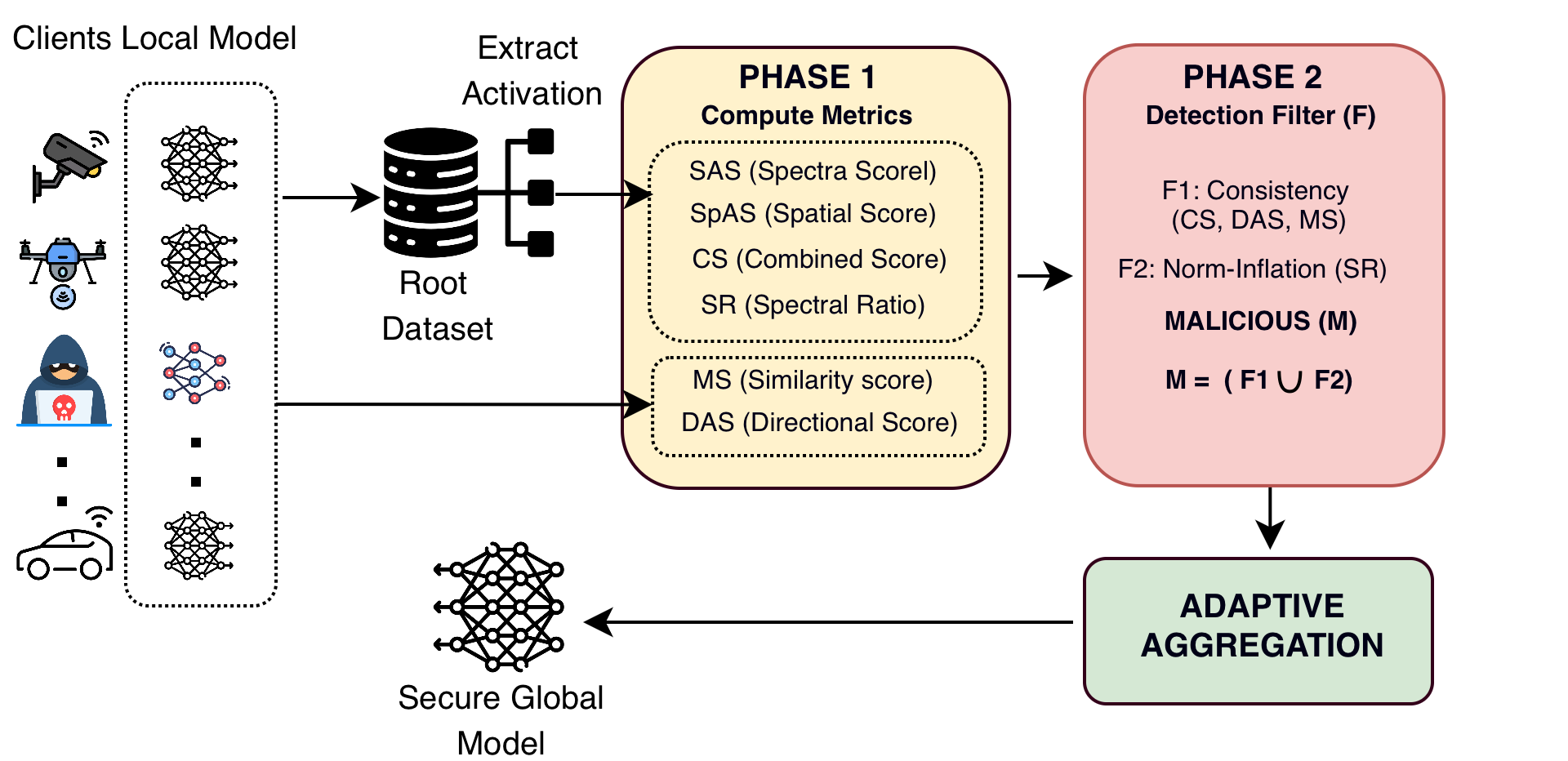}
\caption{High-level overview of FeRA.}
\label{fig:FeRA_main-6}
\end{figure}

We present FeRA, a novel multi-dimensional backdoor detection framework that exploits the fundamental consistency constraint inherent to backdoor attacks. FeRA operates through comprehensive behavioral analysis across six complementary metrics, applying two detection mechanisms that create complementary constraints on adversarial behavior. It executes as a server-side filtering layer between client model reception and global aggregation, analyzing representation-space and parameter-space signatures to identify malicious updates. The defense operates through a two-phase detection pipeline followed by selective aggregation. Fig.~\ref{fig:FeRA_main-6} presents an overview of the methodology.

\textbf{Representation/Activation extraction.} FeRA operates as a pre-aggregation defense and maintains a small root dataset for client representation extraction. For each training round, the global model is initialized and sent to all clients selected for the round. The clients trains the models and sends their local model to the server. Using the root dataset, the server extracts representations from each local models and the global model for that round.

\textbf{Phase 1: Comprehensive metric computation.} For each received client model, FeRA computes six behavioral metrics that capture distinct aspects of backdoor-induced behavioral signatures. The metrics divide into two categories: (1) \textit{representation-based metrics} extracted from learned feature representations, comprising Spectral Attention Score (SAS), Spatial Attention Score (SpAS), their normalized combination, called Combined Score (CS), and (2) \textit{parameter-based metrics} computed from model weight distributions, comprising Directional Attention Score (DAS), Mutual Similarity (MS), and Spectral Ratio (SR) for norm-inflation detection. 

\textbf{Phase 2: Detection filtering.} FeRA applies two detection filters, namely consistency filter and norm-inflation filter as discussed in Section~\ref{sec:observation}, each targeting distinct attack strategies. A client flagged by either mechanism is classified as malicious, establishing comprehensive attack coverage without requiring simultaneous extreme deviations across all dimensions.

\textbf{Adaptive aggregation.} Following detection, FeRA clips the flagged clients and aggregates along with the benign clients using standard FedAvg, computing the weighted average of their model updates based on local dataset sizes. 

\subsection{FeRA in Detail}

This section presents each phase of FeRA's detection pipeline in detail, describing the algorithmic implementation that realizes the framework overviewed in Section~\ref{method}. Alg.~\ref{alg:fera-main} provides the complete FeRA procedure, with detailed subroutines for metric computation (Alg.~\ref{alg:metrics}) and filter application (Alg.~\ref{alg:filters}) presented in their respective subsections.

\begin{algorithm}[t]
\caption{\textsc{FeRA}}
\label{alg:fera-main}
\small
\begin{algorithmic}[1]
\Require Client updates $\{(\boldsymbol{\theta}_i, n_i)\}_{i=1}^N$, global model $\boldsymbol{\theta}_g$, root data $\mathcal{D}_{root}$, clip factor $\beta$
\Ensure Updated global model $\boldsymbol{\theta}^{(t+1)}$
\State $\mathcal{S} \gets \textsc{ComputeMetrics}(\{(\boldsymbol{\theta}_i, n_i)\}, \boldsymbol{\theta}_g, \mathcal{D}_{root})$
\State $\mathcal{M} \gets \textsc{ApplyFilters}(\mathcal{S})$
\For{$i = 1$ to $N$}
    \State $\alpha_i \gets \beta$ if $i \in \mathcal{M}$ else $1$
\EndFor
\State $\boldsymbol{\theta}^{(t+1)} \gets \boldsymbol{\theta}_g + \eta \sum_{i=1}^{N} \frac{n_i}{\sum_j n_j} \alpha_i (\boldsymbol{\theta}_i - \boldsymbol{\theta}_g)$
\Return $\boldsymbol{\theta}^{(t+1)}$
\end{algorithmic}
\end{algorithm}

\subsubsection{Representation/Activation extraction}
FeRA's detection pipeline begins with extracting learned representations from client models, followed by computing six behavioral metrics across two analytical dimensions.
For each client model $f_i^{(t)}$ received at round $t$, we extract feature activations at the penultimate fully-connected layer using a small root dataset $\mathcal{D}_{\text{root}}$ maintained by the server. The root requires no overlap with client training data. Layer selection follows the principle that backdoor signatures manifest most strongly in task-relevant semantic representations immediately preceding classification \cite{tran2018spectral}. We also carried out ablations studies to see the impact of using different layers in Appendix~\ref{app:layer_selection}.

Given sample $(x, y) \in \mathcal{D}_{\text{root}}$, we extract activations $h_\ell^{(i)}(x) = \text{feat}_\ell(f_i^{(t)}, x) \in \mathbb{R}^{d'}$ where $d'$ denotes the layer's output dimension. Extracting representations across all root samples yields feature matrix:
\begin{equation}
\mathbf{H}_i = \begin{bmatrix} h_\ell^{(i)}(x_1)^\top \\ \vdots \\ h_\ell^{(i)}(x_n)^\top \end{bmatrix} \in \mathbb{R}^{n \times d'}
\end{equation}

We similarly extract the global model's representation matrix $\mathbf{H}_{\text{global}}$ using the current global model $f^{(t-1)}$ before aggregation. These representations form the basis for computing representation-based behavioral metrics. The choice of representation dimension $d'$ affects detection sensitivity, as higher-dimensional feature spaces provide richer variance signatures for distinguishing benign and malicious patterns. We investigate the impact of feature dimensionality on detection performance in Appendix~\ref{app:feature_dimension}, demonstrating that FeRA maintains robust detection across a wide range of representation dimensions.

\begin{algorithm}[t]
\caption{\textsc{ComputeMetrics}}
\label{alg:metrics}
\small
\begin{algorithmic}[1]
\Require Client updates $\{(\boldsymbol{\theta}_i, n_i)\}_{i=1}^N$, global model $\boldsymbol{\theta}_g$, root data $\mathcal{D}_{root}$
\Ensure Metric set $\mathcal{S} = \{\sigma,\delta,s^{\text{comb}},\text{DAS},\text{MutualSim},r\}$
\For{$i = 1$ to $N$}
  \State $\mathbf{H}_i \gets \textsc{ExtractFeatures}(\boldsymbol{\theta}_i, \mathcal{D}_r)$
  \State Compute $\sigma_i, \delta_i$ using ~\eqref{eq:spectral} --~\eqref{eq:spatial} \Comment{Spectral \& Spatial}
\EndFor
\State Compute $s_i^{\text{comb}}$ using ~\eqref{eq:combine} \Comment{Combined Score}
\For{$i = 1$ to $N$}
  \State Compute $\text{DAS}_i$ using ~\eqref{eq:DAS} \Comment{Directional Attention}
  \State Compute $\text{MutualSim}_i$ using ~\eqref{eq:mutual} \Comment{Mutual Similarity}
  \State Compute $r_i$ using ~\eqref{eq:spectral-ratio} \Comment{Spectral Ratio}
\EndFor
\State \Return $\mathcal{S}$
\end{algorithmic}
\end{algorithm}

\subsubsection{Phase 1: Metric Computation}

We compute six complementary metrics that capture distinct aspects of representation-space behavior under backdoor attacks. Alg.~\ref{alg:metrics} summarizes the full procedure, showing how these six metrics are derived from client representations and parameters.

\textbf{Spectral attention score (SAS).}
SAS quantifies how representation variance is distributed across principal directions in feature space. For client $i$, we define the representation delta $\Delta_i = H_i - H_{\text{global}}$, center it as $\Delta_{\text{cent}} = \Delta_i - \frac{1}{n}\mathbf{1}_n \mathbf{1}_n^\top \Delta_i$, and compute the covariance $C_{\Delta_i} = \frac{1}{n-1}\Delta_{\text{cent}}^\top \Delta_{\text{cent}}$, which captures the variance structure. We observe that backdoor-induced variance suppression occurs across the dominant spectral subspace rather than a single principal direction: Fig.~\ref{fig:topk_eigenvalues} shows that malicious clients exhibit systematically lower eigenvalues across the top-$k$ principal components ($k{=}5$), indicating multi-dimensional variance collapse. We use the maximum eigenvalue as a computationally efficient summary statistic:
\begin{equation}
\label{eq:spectral}
\sigma_i = \lambda_{\max}(C_{\Delta_i})
\end{equation}

This scalar captures the leading signature of the broader phenomenon: high $\sigma_i$ indicates directional diversity across principal components (benign clients learning from heterogeneous data), while low $\sigma_i$ indicates variance collapse onto low-rank subspaces (backdoor-induced attentional focus on trigger features). The full spectral decomposition $\{\lambda_1, \lambda_2, \ldots, \lambda_k\}$ reveals that malicious clients suppress variance systematically across multiple orthogonal directions, but $\lambda_{\max}$ provides sufficient discriminative power without requiring expensive full eigen-decomposition at scale. This computational efficiency is critical for practical deployment; we analyze FeRA's computational overhead relative to baseline defenses in Appendix~\ref{app:computational}.

\textbf{Spatial attention score (SpAS).} The Spatial Attention Score measures aggregate representation shift magnitude through the Frobenius norm:
\begin{equation}
\label{eq:spatial}
\delta_i = \|\Delta_i\|_F = \sqrt{\sum_{j,k} \Delta_i[j,k]^2}
\end{equation}

While mathematically related to spectral norm through $\|\Delta_i\|_F^2 = \text{tr}(C_{\Delta_i}) = \sum_{k=1}^{d'} \lambda_k$, the Frobenius and spectral norms respond with different sensitivities to eigenvalue distribution shape. Consider two backdoor strategies:

\textit{Directional collapse}: An attack concentrating variance along a single direction produces high $\lambda_{\max}$ (dominant eigenvalue) but potentially moderate $\|\Delta\|_F$ if other dimensions remain active. This pattern evades spectral-only detection while maintaining trigger reliability through consistent principal direction alignment.

\textit{Uniform suppression}: An attack reducing all eigenvalues proportionally produces low $\|\Delta\|_F$ (total displacement drops) while $\lambda_{\max}$ may remain relatively elevated compared to the reduced spectral mass. This pattern evades Frobenius-only detection while achieving trigger consistency through global variance reduction.

High $\delta_i$ indicates substantial representation shifts reflecting natural learning variability from diverse local data. Low $\delta_i$ indicates minimal positional movement, characteristic of backdoor-induced representational stability where poisoned samples consistently map to fixed trigger subspaces. The complementary failure modes of spectral and spatial metrics motivate their joint use in the combined score.

\textbf{Combined score (CS).} To enable unified detection exploiting complementary variance signatures, we normalize both metrics using robust statistics resistant to outliers. For spectral scores $\{\sigma_i\}_{i=1}^N$:
\begin{equation}
\tilde{\sigma}_i = \frac{\sigma_i - \text{median}(\{\sigma_i\})}{\text{IQR}(\{\sigma_i\}) + \epsilon}
\end{equation}
where $\text{IQR}(\cdot) = Q_{75}(\cdot) - Q_{25}(\cdot)$ denotes the interquartile range and $\epsilon = 10^{-12}$ ensures numerical stability when all clients exhibit identical scores. This normalization produces z-scores scaled by interquartile range rather than standard deviation, providing robustness to extreme outliers. Unlike standard z-score normalization using mean and standard deviation, the median and IQR remain stable under minority Byzantine attacks where malicious clients cannot control the median when $|M| < 0.5N$. 
The combined score fuses both dimensions:
\begin{equation}
\label{eq:combine}
s_i^{\text{comb}} = w_\sigma \tilde{\sigma}_i + w_\delta \tilde{\delta}_i
\end{equation}

The combined score defines a detection boundary in normalized $(\tilde{\sigma}, \tilde{\delta})$ space. Because both metrics are z-scored using robust statistics, the combined score represents a weighted linear discriminant. The Consistency Filter flags clients in the lower percentiles of this combined score distribution, corresponding to simultaneous suppression across both spectral and spatial dimensions. While spectral and spatial scores exhibit moderate correlation for benign clients, this correlation breaks for malicious clients attempting metric-specific evasion. Attacks optimizing against spectral detection (maintaining high $\lambda_{\max}$) necessarily reduce total displacement $\|\Delta\|_F$ when concentrating variance directionally, creating negative $\tilde{\delta}_i$ values that expose the attack through the spatial component. Conversely, attacks preserving total displacement through uniform perturbations inflate maximum eigenvalues, triggering detection through elevated $\tilde{\sigma}_i$. The weighted combination with $w_\sigma{=}0.6$ and $w_\delta{=}0.4$ reflects empirically-tuned sensitivity to these complementary evasion strategies, with higher weight on spectral variance capturing the primary backdoor signature while spatial magnitude provides secondary confirmation. We explore various weight combinations in Appendix~\ref{app:threshold_sensitivity}.

\textbf{Directional attention score (DAS).} The Directional Attention Score measures directional alignment between client model parameters and the global model through cosine similarity. For client model parameters $\boldsymbol{\theta}_i$ and global model parameters $\boldsymbol{\theta}_{\text{global}}$, we first flatten all parameter tensors into vectors:
\begin{equation}
\boldsymbol{\theta}_i^{\text{flat}} = \text{flatten}(\boldsymbol{\theta}_i) \in \mathbb{R}^P
\end{equation}
Where $P$ denotes the total parameter count. The cosine similarity quantifies directional alignment:
\begin{equation}
\cos(\boldsymbol{\theta}_i, \boldsymbol{\theta}_{\text{global}}) = \frac{\boldsymbol{\theta}_i^{\text{flat}} \cdot \boldsymbol{\theta}_{\text{global}}^{\text{flat}}}{\|\boldsymbol{\theta}_i^{\text{flat}}\|_2 \cdot \|\boldsymbol{\theta}_{\text{global}}^{\text{flat}}\|_2 + \epsilon}
\end{equation}

To map cosine similarity from $[-1, 1]$ to an interpretable $[0, 1]$ range, we define the Directional Attention Score:
\begin{equation}
\label{eq:DAS}
\text{DAS}_i = \frac{\cos(\boldsymbol{\theta}_i, \boldsymbol{\theta}_{\text{global}}) + 1}{2}
\end{equation}

High $\text{DAS}_i \approx 1$ indicates perfect directional alignment with the global optimization trajectory, characteristic of benign clients pursuing the main task objective. Low $\text{DAS}_i \approx 0$ indicates directional deviation, where client updates point in substantially different directions from global convergence, a signature of backdoor injection that simultaneously optimizes for trigger activation alongside task performance.

\textbf{Mutual similarity.} The Mutual Similarity score captures inter-client behavioral coordination through pairwise parameter similarity. For each client $i$, we compute the maximum cosine similarity to any other client:
\begin{equation}
\label{eq:mutual}
\text{MutualSim}_i = \max_{j \neq i} \cos(\boldsymbol{\theta}_i, \boldsymbol{\theta}_j)
\end{equation}

High mutual similarity among a client subgroup indicates coordinated behavioral patterns characteristic of collusion attacks, where multiple malicious clients converge toward shared backdoor representations. Benign clients training on heterogeneous local data exhibit lower pairwise similarities due to natural distribution differences, while colluding attackers optimizing for identical trigger patterns produce systematically elevated similarity scores.

\textbf{Spectral ratio for Norm-Inflation Detection.} To detect scaling attacks that amplify malicious update magnitudes, we compute 
the spectral ratio comparing each client's representation variance against the 
round-wise median, then apply a robust outlier detection test based on median 
absolute deviation (MAD).

For each client $i$ in round $t$, we compute:
\begin{equation}
\label{eq:spectral-ratio}
r_i = \frac{\sigma_i}{\text{median}\{\sigma_j : j \in [1, N]\}}
\end{equation}

Where $\sigma_i = \lambda_{\max}(\mathbf{C}_{\boldsymbol{\Delta}_i})$ denotes 
the spectral attention score from~\eqref{eq:spectral}, and $N$ is the 
number of clients participating in round $t$. To identify anomalous spectral ratios indicative of norm-inflation attacks, we 
employ the MAD, a robust scale estimator resistant 
to outliers~\cite{leys2013detecting}. A client $i$ is flagged by the Norm-Inflation Filter if:
\begin{equation}
\label{eq:mad-threshold}
r_i > \text{median}\{r_j : j \in [1,N]\} + k \cdot \text{MAD}
\end{equation}

Where $k$ is a sensitivity parameter controlling the detection threshold. Following 
robust outlier detection practices~\cite{leys2013detecting}, we use $k=6$ by default, corresponding to flagging extreme spectral ratios that deviate substantially  from the typical client behavior. This MAD-based approach adapts to each round's empirical distribution while remaining robust under the minority Byzantine assumption,  as malicious clients ($|M| < 0.5N$) cannot control the median.

The six metrics span orthogonal dimensions of backdoor behavior: representation variance structure (Spectral/Spatial), parameter directionality (DAS), coordination patterns (Mutual Similarity), and magnitude manipulation (Spectral Ratio). This multi-dimensional coverage ensures that attacks normalizing specific metrics inevitably expose signatures in complementary dimensions, establishing comprehensive detection surface coverage.

\begin{algorithm}[t]
\caption{\textsc{ApplyFilters}}
\label{alg:filters}
\small
\begin{algorithmic}[1]
\Require Metrics $\mathcal{S}$, thresholds $\tau_{\text{comb}}, \tau_{\text{DAS}}, \tau_{\text{mutual}}, \tau_{\text{spec}}$
\Ensure Malicious client set $\mathcal{M}$
\State $\mathcal{M}_c \gets \emptyset$, $\mathcal{M}_n \gets \emptyset$
\For{$i = 1$ to $N$} \Comment{Consistency Filter}
  \If{$\text{rank}(s_i^{\text{comb}}) \le \tau_{\text{comb}}$ \textbf{and} $\text{rank}(\text{DAS}_i) \le \tau_{\text{DAS}}$ \textbf{and} $\text{rank}(\text{MutualSim}_i) \ge \tau_{\text{mutual}}$}
    \State $\mathcal{M}_c \gets \mathcal{M}_c \cup \{i\}$
  \EndIf
\EndFor
\For{$i = 1$ to $N$} \Comment{Norm-Inflation Filter}
    \State Flag clients using ~\eqref{eq:mad-threshold}
    \State $\mathcal{M}_n \gets \mathcal{M}_n \cup \{i\}$
\EndFor
\State \Return $\mathcal{M} \gets \mathcal{M}_c \cup \mathcal{M}_n$
\end{algorithmic}
\end{algorithm}

\subsubsection{Phase 2: Detection Mechanisms}

FeRA applies two complementary detection mechanisms, each targeting distinct attack strategies. Alg.~\ref{alg:filters} formalizes the detection procedure, showing how both filters operate on the computed metrics to identify malicious clients.

\textbf{Consistency filter.} This filter identifies attacks that prioritize reliability through suppressed variance and temporal stability. The filter examines clients across three dimensions simultaneously: combined representation score (capturing variance suppression), directional attention (measuring global alignment), and mutual similarity (detecting peer coordination). Rather than using fixed thresholds, the filter employs percentile-based ranking adapted to each round's empirical distribution. It ranks all clients by three metrics: their representation variance (combined spectral/spatial score); directional alignment with the global model (DAS); and mutual similarity. It then defines decision surfaces based on the empirical tails of these distributions. A client is flagged if it falls in the lower quantile region for variance and alignment metrics while simultaneously residing in the upper quantile for similarity to peers. This combination captures the fundamental trade-off where maintaining attack reliability requires behavioral consistency that manifests as suppressed representation diversity, while the mutual similarity component captures coordination patterns that emerge when colluding attackers optimize identical backdoor objectives despite data heterogeneity. This rank-based approach adapts to the observed client behavior every round, eschewing any hardcoded cut-off.

\textbf{Norm-Inflation Filter.} The Norm-Inflation Filter identifies clients with 
anomalously inflated spectral ratios using a MAD-based robust outlier test. We compute each client's spectral ratio $r_i$ (as in~\eqref{eq:spectral-ratio}) 
and the round-wise MAD. Clients whose spectral ratio 
exceeds the median by more than $k$ MADs are 
flagged as exhibiting norm-inflation behavior. Scaling attacks deliberately inflate 
update magnitudes to dominate aggregation, producing spectral ratios exceeding those 
of benign clients (as illustrated in Fig.~\ref{fig:norm_inflation}). The MAD-based 
threshold adapts to the scale and variability of each round's client behavior, 
providing robustness to benign heterogeneity while exposing magnitude-manipulation 
strategies. Following best practices in robust outlier detection~\cite{leys2013detecting}, 
we set $k=6$ by default, though ablation studies (Appendix~\ref{app:threshold_sensitivity}) 
demonstrate stable performance across $k \in [3, 10]$.

\subsubsection{Adaptive Aggregation} Following detection, FeRA applies a graduated aggregation strategy that balances security and utility preservation. Rather than completely excluding flagged clients, which risks accuracy degradation from false positives under severe data heterogeneity, we employ \textit{adaptive update clipping}. Let $\mathcal{M} = \mathcal{M}_{\text{consistency}} \cup \mathcal{M}_{\text{norm}}$ denote the union of clients flagged by either detection mechanism. The aggregation proceeds as:
\begin{equation}
\boldsymbol{\theta}^{(t+1)} = \boldsymbol{\theta}^{(t)} + \eta \sum_{i=1}^{N} \frac{|D_i|}{|D|} \cdot \alpha_i \cdot \boldsymbol{\Delta}_i^{(t)}
\end{equation}
where the clipping factor is defined as $\alpha_i = \beta$ for $i \in \mathcal{M}$ and $\alpha_i = 1$ otherwise.

with $\beta \in [0, 1]$ controlling the contribution of flagged clients (default: $\beta = 0.1$). This graduated trust mechanism ensures that: (1) truly malicious clients have their backdoor contributions suppressed by a factor of $1/\beta = 10\times$, rendering backdoor injection ineffective, while (2) benign clients incorrectly flagged due to extreme non-IID distributions or natural variance still contribute to model convergence, preventing accuracy degradation. The clipping approach addresses the challenge of false positive accumulation (C3) by providing a safety margin: even if adaptive attackers occasionally evade detection, their unclipped contributions remain bounded by the minority assumption, while detected attacks are sufficiently attenuated to prevent backdoor establishment.

Specific quantile thresholds used during evaluation are treated as hyper-parameters and detailed in Section~\ref{sec:setup}.

\begin{table*}[t]
\centering
\caption{Performance comparison on IID CIFAR-10 under diverse backdoor attacks. Attackers begin poisoning at round 2001 for 100 rounds. Results reported as MA(\%) / BA(\%). Bold indicates best result in each category (highest MA, lowest BA).}
\label{tab:iid_main}
\resizebox{\textwidth}{!}{
\begin{tabular}{l|c|cccccc|cc}
\toprule
\textbf{Defense} & \textbf{No Attack} & \textbf{BadNet~\cite{gu2019badnets}} & \textbf{Blended~\cite{chen2017targeted}} & \textbf{Edge-case~\cite{wang2020attack}} & \textbf{IBA~\cite{nguyen2023iba}} & \textbf{Neurotoxin~\cite{zhang2022neurotoxin}} & \textbf{Chameleon~\cite{dai2023chameleon}} & \textbf{Avg. MA} & \textbf{Avg. BA} \\
\midrule
No Defense & 88.71 / 0.00 & 88.23 / 94.35 & 87.92 / 92.18 & 88.14 / 89.76 & 87.58 / 93.81 & 86.82 / 95.24 & 87.36 / 91.67 & 87.79 & 92.78 \\
\midrule
Multi-Krum~\cite{blanchard2017machine} & 87.12 / 0.00 & 76.34 / 88.42 & 74.85 / 85.73 & 75.91 / 84.26 & 76.84 / 87.35 & 73.27 / 89.18 & 76.83 / 86.94 & 76.69 & 86.93 \\
FoolsGold~\cite{fung2020foolsgold} & 86.47 / 0.00 & 82.16 / 76.38 & 80.72 / 72.85 & 81.34 / 74.12 & 79.85 / 78.74 & 79.63 / 78.46 & 81.95 / 75.79 & 81.11 & 76.03 \\
FLAME~\cite{nguyen2022flame} & 87.94 / 0.00 & 85.26 / 12.47 & 84.15 / 15.28 & 84.63 / 13.84 & 83.75 / 16.42 & 82.91 / 18.35 & 84.78 / 14.16 & 84.42 & 15.07 \\
FLTrust~\cite{cao2020fltrust} & 88.35 / 0.00 & 86.74 / 8.26 & 85.48 / 11.53 & 86.16 / 9.74 & 84.93 / 12.85 & 84.82 / 13.27 & 86.24 / 10.48 & 85.73 & 11.02 \\
RLR~\cite{ozdayi2021defending} & 79.26 / 0.00 & 76.24 / 2.18 & 77.93 / 3.46 & 77.75 / 2.84 & 73.64 / \textbf{4.57} & 78.26 / 4.23 & 77.94 / 3.47 & 76.68 & 3.46 \\
DeepSight~\cite{rieger2022deepsight} & 85.94 / 0.00 & 81.75 / 45.26 & 80.34 / 48.73 & 81.92 / 46.85 & 79.27 / 51.34 & 79.46 / 52.18 & 82.15 / 47.36 & 81.05 & 48.64 \\
\midrule
\textbf{FeRA (Ours)} & \textbf{88.42 / 0.00} & \textbf{87.93} / \textbf{1.26} & \textbf{87.24} / \textbf{3.35} & \textbf{87.63} / \textbf{0.94} & \textbf{86.95} / 5.08 & \textbf{86.54} / \textbf{0.93} & \textbf{87.85} / \textbf{1.47} & \textbf{87.42} & \textbf{2.17} \\
\bottomrule
\end{tabular}
}
\end{table*}
\section{Experimental Evaluation}
\label{sec:experiments}

In this section we conduct extensive experiments to evaluate FeRA's detection performance across diverse attack scenarios, datasets, and system configurations.

\subsection{Experimental Setup}
\label{sec:setup}

\textbf{Datasets and models.} We evaluate FeRA on four widely-used image classification benchmarks: CIFAR-10, CIFAR-100~\cite{krizhevsky2009learning}, Tiny-ImageNet, and German Traffic Sign Recognition Benchmark (GTSRB) \cite{GTSRB2012Zenodo} dataset. For model architectures, we employ ResNet-18 \cite{He2016DeepResidual} as our primary architecture, with additional evaluation on ResNet-34 \cite{He2016DeepResidual} and VGG-16 \cite{Simonyan2014VeryDeep} to demonstrate architecture-agnostic effectiveness.

\textbf{FL configuration.} Our FL setup uses $N=100$ clients, with $m=10$ sampled uniformly at random each round to train the global model. A fixed subset of clients is set to be malicious (by default $10\%$ of the population, unless stated otherwise), and their identities remain constant throughout training. Each round may include zero, one, or multiple malicious clients, with their proportion in the sampled set capped at $40\%$.

We consider both IID and non-IID data partitions. For the IID case, training data are evenly distributed across clients. For the non-IID case, we follow the standard Dirichlet procedure~\cite{hsu2019measuring,huang2023multimetrics,huang2023lockdown}, drawing each client’s data proportion from $\text{Dir}(\alpha)$ with default $\alpha = 0.5$. Local training uses SGD with server learning rate $\eta = 0.5$, client learning rate $\alpha = 0.1$, and two local epochs per round. As is common in FL, we use dataset-specific pretraining before enabling attacks: CIFAR-10 is pretrained for 2000 rounds, CIFAR-100 for 1000, and Tiny-ImageNet for 500; GTSRB and MNIST variants are trained from scratch, with attacks starting after 50 and 5 rounds, respectively. Unless otherwise specified, attacks are then run for 100 rounds. The server aggregates updates using FedAvg~\cite{mcmahan2017communication}.

FeRA employs percentile-based detection thresholds. The consistency filter uses quantile thresholds $\tau_{\text{comb}} = 0.50$ (combined score), $\tau_{\text{DAS}} = 0.50$ (DAS), and $\tau_{\text{mutual}} = 0.60$ (mutual similarity). The combined representation-variance score uses weights $w_{\text{spectral}} = 0.6$ and $w_{\delta} = 0.4$. The Norm-Inflation Filter applies MAD-based outlier detection with sensitivity parameter $k = 6$. All threshold and weighting parameters are treated as tunable hyperparameters; Appendix~\ref{app:threshold_sensitivity} reports ablations on their sensitivity.

\textbf{Attack scenarios.} We evaluate against eight attack variants: BadNet~\cite{gu2019badnets}, Blended~\cite{chen2017targeted}, Edge-case~\cite{wang2020attack}, DBA~\cite{xie2019dba}, Neurotoxin~\cite{zhang2022neurotoxin},Chameleon~\cite{dai2023chameleon}, IBA \cite{nguyen2023iba}, A3FL~\cite{zhang2023a3fl}, and an adaptive attack. In our ablation studies we also evaluate against Anticipate \cite{wen2022thinking} attack which by principle of its design can evade some of our defense filter. During training, attackers may control single or multiple clients, initiating poisoning from the poisoning starting round as described initially.

\textbf{Baseline defenses.} We compare FeRA with the non-robust baseline FedAvg and against state-of-the-art defenses: Multi-Krum~\cite{blanchard2017machine}, FoolsGold~\cite{fung2020foolsgold}, FLAME~\cite{nguyen2022flame}, FLTrust~\cite{cao2020fltrust}, RLR~\cite{ozdayi2021defending}, Lockdown \cite{huang2023lockdown} and Deepsight~\cite{rieger2022deepsight}.

\textbf{Evaluation metrics.} Performance is assessed using \textit{Main task Accuracy (MA)}, measuring classification performance on benign test data; and \textit{Backdoor Accuracy (BA)}, measuring attack success rate when triggers are present. Lower BA with minimal MA degradation indicates superior defense.

\subsection{Performance on IID Data}

Table~\ref{tab:iid_main} presents results under IID CIFAR-10 settings across six attack types. FeRA achieves 2.17\% average BA while maintaining 87.42\% average MA, outperforming all baselines. Against BadNet, FeRA limits BA to 1.26\% at 87.93\% MA, whereas competing methods either fail to mitigate attacks (Multi-Krum: 88.42\% BA, FoolsGold: 76.38\% BA, Deepsight: 45.26\% BA) or sacrifice utility (FLAME: 12.47\% BA at 85.26\% MA). For Neurotoxin, which achieves 95.24\% BA without defense, FeRA reduces BA to 0.93\% while maintaining 86.54\% MA.

FeRA provides 80.29\% BA reduction compared to FLTrust (11.02\% average BA) while suffering only 0.58\%-1.69\% MA degradation versus no-defense baselines. RLR achieves comparable BA (2.18\%-4.57\%) but incurs 17.42\% MA loss (76.24\%-73.64\% MA). BA variance across attacks is 4.15\% for FeRA versus 5.88\% for FLAME and 5.01\% for FLTrust, confirming consistent detection regardless of trigger type or injection strategy.

\begin{table}[t]
\centering
\caption{Performance on CIFAR-100 and Tiny-ImageNet under pixel-pattern attacks. Poisoning begins at round 2001 for 100 rounds. Results: MA(\%) / BA(\%). Bold indicates best result in each column.}
\label{tab:other_datasets}
\resizebox{\columnwidth}{!}{
\begin{tabular}{l|cc|cc}
\toprule
\textbf{Defense} & \textbf{CIFAR-100} & \textbf{Tiny-ImageNet} & \textbf{Avg. MA} & \textbf{Avg. BA} \\
\midrule
No Defense & 62.84 / 84.57 & 45.36 / 88.25 & 54.10 & 86.41 \\
\midrule
Multi-Krum~\cite{blanchard2017machine} & 58.47 / 87.85 & 41.73 / 89.74 & 50.10 & 88.80 \\
FoolsGold~\cite{fung2020foolsgold} & 61.25 / 84.46 & 43.84 / 86.57 & 52.55 & 85.52 \\
FLAME~\cite{nguyen2022flame} & 59.74 / 12.85 & 42.16 / 18.47 & 50.95 & 15.66 \\
FLTrust~\cite{cao2020fltrust} & 60.46 / 9.47 & 43.64 / 12.75 & 52.05 & 11.11 \\
RLR~\cite{ozdayi2021defending} & 44.35 / 0.46 & 32.74 / 0.42 & 38.55 & 0.44 \\
DeepSight~\cite{rieger2022deepsight} & 57.94 / 61.35 & 40.26 / 64.83 & 49.10 & 63.09 \\
\midrule
\textbf{FeRA (Ours)} & \textbf{61.84} / \textbf{0.35} & \textbf{44.16} / \textbf{0.12} & \textbf{53.00} & \textbf{0.24} \\
\bottomrule
\end{tabular}
}
\end{table}

\subsection{Performance on Additional Datasets}
\textbf{Generalization across dataset complexities.} To validate FeRA's robustness beyond CIFAR-10, we extend our evaluation to CIFAR-100 (100 classes, finer-grained categories) and Tiny-ImageNet (200 classes, more complex visual patterns). Table~\ref{tab:other_datasets} demonstrate FeRA's consistent effectiveness across varying dataset complexities.

\textbf{Dataset complexity robustness.} FeRA maintains strong performance on complex datasets with numerous classes (CIFAR-100: 100 classes) and high-resolution images (Tiny-ImageNet: 64×64, 200 classes). Average BA of 0.24\% represents 97.84\% reduction compared to FLTrust (11.11\%) and 98.47\% reduction compared to FLAME (15.66\%). On CIFAR-100, FeRA achieves 0.35\% BA with only 1.00\% MA degradation, whereas FLTrust incurs 2.38\% MA loss while achieving 9.47\% BA. For Tiny-ImageNet, FeRA limits BA to 0.12\% with 44.16\% MA. Other methods either fail completely (Multi-Krum: 89.74\% BA, FoolsGold: 86.57\% BA, Deepsight: 64.83\% BA) or sacrifice substantial utility like RLR with 32.74\% MA. FeRA's effectiveness on Tiny-ImageNet shows that representation-level variance suppression remains detectable even when task complexity increases.

\subsection{Resilience Under Data Heterogeneity}
% \textcolor{red}{Check Table 3 results. FeRA's MA is not the best one in some categories. Or perhaps you say Bold indicates the best MA/BA combination?}
Data heterogeneity poses a fundamental challenge for backdoor defense, as non-IID distributions amplify natural variation in client updates, potentially masking malicious patterns. Table~\ref{tab:non_iid} examines FeRA's robustness across varying non-IID degrees ($\alpha \in \{0.2, 0.5, 0.7\}$) for Neurotoxin attacks on CIFAR-10. FeRA maintains consistently low BA (1.35\%--1.93\%) across all heterogeneity levels, with 0.58\% variance. At extreme heterogeneity ($\alpha=0.2$), FeRA achieves 1.74\% BA with 84.05\% MA, compared to Multi-Krum (89.26\% BA), FoolsGold (68.47\% BA), and Deepsight (47.84\% BA). FLAME and FLTrust achieve 14.73\% and 10.35\% BA respectively; FeRA provides 88.18\% and 83.19\% reductions. As $\alpha$ increases from 0.2 to 0.7, FLAME's BA decreases from 14.73\% to 9.26\% and FLTrust from 10.35\% to 7.16\%, reflecting that distance-based and trust-score methods benefit from reduced benign update variance under more IID conditions. FeRA's BA remains stable at 1.67\% average, confirming that representation-level variance suppression persists regardless of data distribution characteristics.

FeRA maintains 84.05\%--87.16\% MA across all $\alpha$ values. At $\alpha=0.2$, FeRA achieves 84.05\% MA, only 4.21\% below the no-defense baseline (88.26\% MA), with degradation primarily from non-IID learning difficulty rather than defense-induced removals. At $\alpha=0.5$, FeRA achieves 87.16\% MA, within 0.69\% of baseline.

\begin{table}[t]
\centering
\caption{Performance under non-IID data distributions. Neurotoxin attack on CIFAR-10, poisoning from round 2001 for 100 rounds. Dirichlet parameter $\alpha$ controls heterogeneity (lower = more severe). Results: MA(\%) / BA(\%). Bold indicates best result in each column.}
\label{tab:non_iid}
\resizebox{\columnwidth}{!}{
\begin{tabular}{l|ccc|cc}
\toprule
\textbf{Defense} & $\alpha$\textbf{=0.2} & $\alpha$\textbf{=0.5} & $\alpha$\textbf{=0.7} & \textbf{Avg. MA} & \textbf{Avg. BA} \\
\midrule
No Defense & 88.26 / 93.74 & 87.85 / 91.47 & 87.53 / 89.84 & 87.88 & 91.68 \\
\midrule
Multi-Krum~\cite{blanchard2017machine} & 76.14 / 89.26 & 77.45 / 87.63 & 78.35 / 85.47 & 77.31 & 87.45 \\
FoolsGold~\cite{fung2020foolsgold} & 81.74 / 68.47 & 82.95 / 72.16 & 83.84 / 76.85 & 82.84 & 72.49 \\
FLAME~\cite{nguyen2022flame} & 84.85 / 14.73 & 85.36 / 11.84 & 85.74 / \ 9.26 & 85.32 & 11.94 \\
FLTrust~\cite{cao2020fltrust} & \textbf{85.94} / 10.35 & 
\ 86.47 / 8.74 & \ 86.85 
/ 7.16 & \textbf{86.42} & \ 8.75 \\
DeepSight~\cite{rieger2022deepsight} & 80.35 / 47.84 & 81.63 / 44.26 & 82.74 / 38.95 & 81.57 & 43.68 \\
\midrule
\textbf{FeRA (Ours)} & \ 84.05 / \textbf{1.74} & \textbf{87.16} / \textbf{1.35} & \textbf{86.94} / \textbf{1.93} & \ 86.05 & \textbf{1.67} \\
\bottomrule
\end{tabular}
}
\end{table}

\begin{table*}[t]
\centering
\caption{Performance of FeRA under varying non-IID severity ($\alpha$) across eight backdoor attacks on CIFAR-10. FeRA maintains low BA under moderate heterogeneity and degrades only under extreme data skew ($\alpha=0.1$), while preserving high MA across all settings. Bold indicates mean result.}
\label{tab:iid_sensitivity}
\begin{tabular}{@{}l|cccccccccc|cc@{}}
\toprule
\multirow{2}{*}{\textbf{Attack}} &
\multicolumn{2}{c}{\textbf{$\alpha=0.1$}} &
\multicolumn{2}{c}{\textbf{$\alpha=0.2$}} &
\multicolumn{2}{c}{\textbf{$\alpha=0.5$}} &
\multicolumn{2}{c}{\textbf{$\alpha=0.7$}} &
\multicolumn{2}{c}{\textbf{$\alpha=1.0$ (IID)}} &
\multirow{2}{*}{\textbf{Avg. MA}} &
\multirow{2}{*}{\textbf{Avg. BA}} \\
\cmidrule(lr){2-11}
& \textbf{MA} & \textbf{BA}
& \textbf{MA} & \textbf{BA}
& \textbf{MA} & \textbf{BA}
& \textbf{MA} & \textbf{BA}
& \textbf{MA} & \textbf{BA} & & \\
\midrule
No attack     & 80.15 & 0.00  & 85.26 & 0.00 & 91.47 & 0.00 & 91.84 & 0.00 & 92.16 & 0.00 & 88.18 & 0.00 \\
\midrule
Pattern~\cite{gu2019badnets}       & 72.04 & 1.85  & 85.47 & 1.16 & 91.26 & 1.53 & 91.35 & 1.14 & 91.84 & 1.36 & 86.39 & 1.41 \\
Pixel~\cite{gu2019badnets}         & 82.47 & 5.63  & 85.05 & 0.74 & 90.05 & 0.94 & 90.26 & 1.63 & 91.85 & 1.26 & 87.94 & 2.04 \\
BadNet~\cite{gu2019badnets}        & 80.16 & 15.74 & 84.36 & 0.84 & 91.63 & 1.26 & 90.63 & 1.74 & 91.74 & 1.36 & 87.70 & 4.19 \\
Blended~\cite{chen2017targeted}       & 79.94 & 18.69 & 86.74 & 4.05 & 91.36 & 1.53 & 89.74 & 5.90 & 91.47 & 1.14 & 87.85 & 6.26 \\
Edge-case~\cite{wang2020attack}     & 82.85 & 2.05  & 83.84 & 0.63 & 90.94 & 1.36 & 88.05 & 4.84 & 92.05 & 0.94 & 87.55 & 1.96 \\
A3FL~\cite{zhang2023a3fl}          & 80.16 & 2.05  & 86.74 & 0.63 & 89.63 & 1.05 & 91.53 & 1.26 & 91.74 & 1.26 & 87.96 & 1.25 \\
IBA~\cite{nguyen2023iba}           & 79.16 & 12.83 & 82.47 & 6.25 & 91.16 & 5.13 & 90.47 & 4.12 & 91.94 & 2.94 & 87.04 & 6.25 \\
DBA~\cite{xie2019dba}              & 81.63 & 2.16  & 86.74 & 1.05 & 90.63 & 0.84 & 91.26 & 1.36 & 91.84 & 1.47 & 88.42 & 1.38 \\
\midrule
\textbf{Mean} & \textbf{79.80} & \textbf{7.63} & \textbf{84.95} & \textbf{1.92} & \textbf{90.83} & \textbf{1.70} & \textbf{90.41} & \textbf{2.75} & \textbf{91.81} & \textbf{1.47} & \textbf{87.61} & \textbf{3.09} \\
\bottomrule
\end{tabular}
\end{table*}

\textbf{Robustness across data heterogeneity levels on more attacks.} Table~\ref{tab:iid_sensitivity} evaluates FeRA on CIFAR-10 under varying non-IID severity ($\alpha \!\in\! \{0.1, 0.2, 0.5, 0.7, 1.0\}$) across eight backdoor variants. FeRA sustains low mean BA at moderate heterogeneity, achieving 1.92\% at $\alpha=0.2$ and 1.70\% at $\alpha=0.5$. Under extreme non-IID conditions ($\alpha=0.1$), the mean BA increases to 7.63\%, driven mainly by three attacks (BadNet: 15.74\%, Blended: 18.69\%, IBA: 12.83\%). Pattern-based attacks (Pattern, Pixel, DBA, A3FL, Edge-case) remain below 2.16\% across all $\alpha$, indicating that variance-based signals remain detectable even when client distributions diverge. MA increases steadily from 79.80\% at $\alpha=0.1$ to 91.81\% under IID data, remaining within 0.30\% of no-defense baselines.
\subsection{Cross-Silo Federated Learning}

\textbf{Effectiveness in cross-silo setting.} To evaluate FeRA in a different FL deployment scenario, we simulate cross-silo FL with 10 total clients where all clients participate each round. Unlike the cross-device setting in Section 5.1 (100 clients with 10 sampled per round), this cross-silo configuration models organizational FL where a small number of institutions (e.g., hospitals, banks) collaborate with full participation. This setting presents distinct challenges: with fewer clients and full participation, malicious clients constitute a higher proportion of each round's updates, potentially overwhelming detection mechanisms. Results in Table~\ref{tab:cross_device} validate FeRA's robustness under this more adversarial deployment constraint.

\begin{table}[t]
\centering
\caption{Performance in cross-silo FL with 10 total clients, all participating each round. Pixel-pattern attack on CIFAR-10 under IID and non-IID ($\alpha=0.5$) settings. Results: MA(\%) / BA(\%). Bold indicates best result in each column.}
\label{tab:cross_device}
\resizebox{\columnwidth}{!}{
\begin{tabular}{l|cc|cc}
\toprule
\textbf{Defense} & \textbf{IID} & \textbf{Non-IID} & \textbf{Avg. MA} & \textbf{Avg. BA} \\
\midrule
No Defense & 88.36 / 92.84 & 85.74 / 91.47 & 87.05 & 92.16 \\
\midrule
FoolsGold~\cite{fung2020foolsgold} & 88.25 / 69.16 & 85.73 / 20.25 & 86.99 & 44.71 \\
Lockdown~\cite{huang2023lockdown} & 88.14 / 6.35 & 81.36 / 10.84 & 84.75 & 8.60 \\
FLAME~\cite{nguyen2022flame} & 83.94 / 18.74 & 81.26 / 22.47 & 82.60 & 20.61 \\
FLTrust~\cite{cao2020fltrust} & 85.16 / 14.36 & 82.84 / 17.95 & 84.00 & 16.16 \\
RLR~\cite{ozdayi2021defending} & 79.26 / 2.35 & 59.94 / 3.36 & 69.60 & 2.86 \\
\midrule
\textbf{FeRA (Ours)} & \textbf{88.25} / \textbf{0.94} & \textbf{85.74} / \textbf{1.93} & \textbf{87.00} & \textbf{1.44} \\
\bottomrule
\end{tabular}
}
\end{table}

FeRA achieves 0.94\% BA under IID and 1.93\% under non-IID cross-silo settings, demonstrating robustness to client sampling dynamics. FoolsGold failed (BA$>60\%$), Lockdown and RLR show improved performance (6.35\% and 2.35\% BA respectively) in IID setting. The minimal gap between IID (0.94\% BA) and non-IID (1.93\% BA) cross-silo settings validates FeRA's adaptability. FeRA maintains 88.25\% and 85.74\% MA respectively, representing only 0.11\% and 0\% degradation from no-defense baselines. This confirms that FeRA's representation-level detection remains effective regardless of sampling patterns or data heterogeneity, making it suitable for realistic large-scale FL deployments where client participation is intermittent and unpredictable.

\subsection{Ablation Studies}
\label{ablation}

To understand how each detection mechanism contributes to FeRA's overall effectiveness, we conduct systematic ablation studies on Neurotoxin and Anticipate attacks under non-IID settings ($\alpha=0.5$). By isolating the Consistency Filter and Norm-Inflation Filter, we reveal their complementary roles in addressing different attack strategies. Results are presented in Table~\ref{tab:ablation}.

\begin{table}[t]
\centering
\caption{Ablation study: detection mechanism contributions under two attack scenarios. CIFAR-10, $\alpha=0.5$, poisoning from round 2001 for 100 rounds. Results: MA(\%) / BA(\%).}
\label{tab:ablation}
\resizebox{\columnwidth}{!}{
\begin{tabular}{lcc}
\toprule
\textbf{Configuration} & \textbf{Neurotoxin~\cite{zhang2022neurotoxin}} & \textbf{Anticipate~\cite{wen2022thinking}} \\
\midrule
No Defense & 90.16 / 95.24 & 87.15 / 93.74 \\
\midrule
Consistency Filter Only & 87.26 / 1.74 & 85.05 / 95.53 \\
+ Norm-Inflation Filter & 87.16 / 1.35 & 91.05 / 1.84 \\
%\midrule
\bottomrule
\end{tabular}
}
\end{table}

\textbf{Complementary detection mechanisms.} The two detection mechanisms address distinct attack strategies with minimal overlap. For Neurotoxin attacks, the Consistency Filter alone reduces BA from 95.24\% to 1.74\% (98.17\% attack suppression), demonstrating that representation-level variance suppression combined with coordination pattern analysis provides robust detection. Adding the Norm-Inflation Filter marginally improves BA to 1.35\% while slightly reducing MA to 87.16\%, as Neurotoxin primarily relies on stealth rather than magnitude manipulation. The Anticipate attack \cite{wen2022thinking} reveals the complementary nature of these mechanisms. Anticipate optimizes poisoning by strategically anticipating future client behaviors, generating persistent backdoors through careful update planning that deliberately injects variance to evade consistency-based detection. With the Consistency Filter alone, BA remains at 95.53\%, as the attack successfully mimics benign variance patterns. However, adding the Norm-Inflation Filter reduces BA to 1.84\% with 91.05\% MA, exposing Anticipate's reliance on norm manipulation to maintain backdoor effectiveness while evading variance-based detection.

\textbf{Root dataset size sensitivity.} Table~\ref{tab:root_size} investigates root dataset size requirements for optimal detection performance on CIFAR-10 with Neurotoxin attacks. Performance stabilizes at $|D_{\text{root}}|=64$, achieving 1.35\% BA with 87.26\% MA. Smaller datasets ($|D_{\text{root}}|=32$) yield 3.63\% BA due to insufficient representation diversity, with limited samples, representation statistics lack reliability. Larger datasets maintain similar performance, confirming 64 samples provide sufficient statistical power. Table~\ref{tab:root_size_ood} evaluates FeRA using CIFAR-100 samples (100 classes, different objects) as root dataset for CIFAR-10 detection. Even with this severe distribution mismatch, FeRA achieves 3.64\% BA at $|D_{\text{root}}|=64$, with only modest MA degradation (84.47\% vs 87.26\% for in-distribution root). This demonstrates FeRA's robustness to root dataset selection: even public datasets from different domains suffice for effective detection.

\begin{table}[t]
\centering
\caption{Root dataset size sensitivity analysis. Neurotoxin attack on CIFAR-10, $\alpha=0.5$, poisoning from round 2001 for 100 rounds. Results: MA(\%) / BA(\%).}
\label{tab:root_size}
\resizebox{0.8\columnwidth}{!}{
\begin{tabular}{lccccc}
\toprule
\textbf{$|D_{\text{root}}|$} & \textbf{32} & \textbf{64} & \textbf{128} & \textbf{256} & \textbf{512} \\
\midrule
MA (\%) & 85.84 & 87.26 & 89.15 & 90.53 & 90.16 \\
BA (\%) & 3.63 & 1.35 & 1.16 & 1.05 & 0.74 \\
\bottomrule
\end{tabular}
}
\end{table} 

\begin{table}[t]
\centering
\caption{Root dataset size sensitivity analysis. Neurotoxin attack on CIFAR-10 with CIFAR100 as OOD dataset, $\alpha=0.5$, poisoning from round 2001 for 100 rounds. Results: MA(\%) / BA(\%).}
\label{tab:root_size_ood}
\resizebox{0.5\columnwidth}{!}{
\begin{tabular}{lccc}
\toprule
\textbf{$|D_{\text{root}}|$} & \textbf{32} & \textbf{64} & \textbf{128} \\
\midrule
MA (\%) & 81.85 & 84.47 & 86.16 \\
BA (\%) & 4.15 & 3.64 & 2.63 \\
\bottomrule
\end{tabular}
}
\end{table}
\subsection{Architecture Generalization}

\textbf{Performance across network architectures.} Defense mechanisms that rely on architecture-specific properties may fail when deployed on different model families. We evaluate FeRA's generalization across CNN variants with different depths and architectural designs to assess its applicability beyond the default ResNet-18. Table~\ref{tab:architectures} demonstrates FeRA's architecture-agnostic effectiveness.

\begin{table}[t]
\centering
\caption{Performance across model architectures on CIFAR-10. Pixel-pattern attack, poisoning from round 1200 for 250 rounds. Results: MA(\%) / BA(\%). Bold indicates best result in each column.}
\label{tab:architectures}
\resizebox{\columnwidth}{!}{
\begin{tabular}{l|ccc|cc}
\toprule
\textbf{Defense} & \textbf{ResNet-18} & \textbf{ResNet-34} & \textbf{VGG-16} & \textbf{Avg. MA} & \textbf{Avg. BA} \\
\midrule
No Defense & 88.26 / 94.35 & 87.84 / 92.74 & 86.94 / 93.85 & 87.68 & 93.65 \\
\midrule
Multi-Krum~\cite{blanchard2017machine} & 76.35 / 88.47 & 75.74 / 86.94 & 74.26 / 87.53 & 75.45 & 87.65 \\
FLAME~\cite{nguyen2022flame} & 85.26 / 12.47 & 84.63 / 14.36 & 83.94 / 13.74 & 84.61 & 13.52 \\
FLTrust~\cite{cao2020fltrust} & 86.74 / 8.26 & 86.16 / 9.84 & 85.36 / 10.47 & 86.09 & 9.52 \\
\midrule
\textbf{FeRA (Ours)} & \textbf{87.94} / \textbf{0.94} & \textbf{87.16} / \textbf{1.26} & \textbf{86.47} / \textbf{1.47} & \textbf{87.19} & \textbf{1.22} \\
\bottomrule
\end{tabular}
}
\end{table}

\textbf{Cross-Architecture Performance.} FeRA maintains BA below 1.50\% across ResNet-18 (0.94\%), ResNet-34 (1.26\%), and VGG-16 (1.47\%), with standard deviation of 0.27\%. This represents 85.96\%--88.62\% BA reduction compared to FLTrust (8.26\%--10.47\%), confirming FeRA's detection principles generalize across architectural choices. The minimal variance demonstrates reliability regardless of network depth (18 vs 34 layers) or design patterns (residual vs sequential).

\subsection{Advanced Attack Scenarios}

Table~\ref{tab:a3fl_cifar10} evaluates FeRA against A3FL~\cite{zhang2023a3fl}, a state-of-the-art trigger-optimization attack. FeRA attains $1.26\%$ BA and $84.26\%$ MA under A3FL, corresponding to a $98.64\%$ BA reduction relative to undefended FedAvg ($92.47\%$ BA). Despite A3FL’s optimization for stealth and minimal parameter perturbation, backdoor samples remain fundamentally out-of-distribution (OOD) with respect to the benign target-class distribution; this OOD nature induces representation-level variance suppression that FeRA exploits. In contrast, DeepSight and RLR fail completely (BA $91.85\%$ and $92.36\%$, respectively), highlighting FeRA’s unique effectiveness against trigger-optimized attacks.

\begin{figure}[t]
    \centering
    \includegraphics[width=0.7\columnwidth]{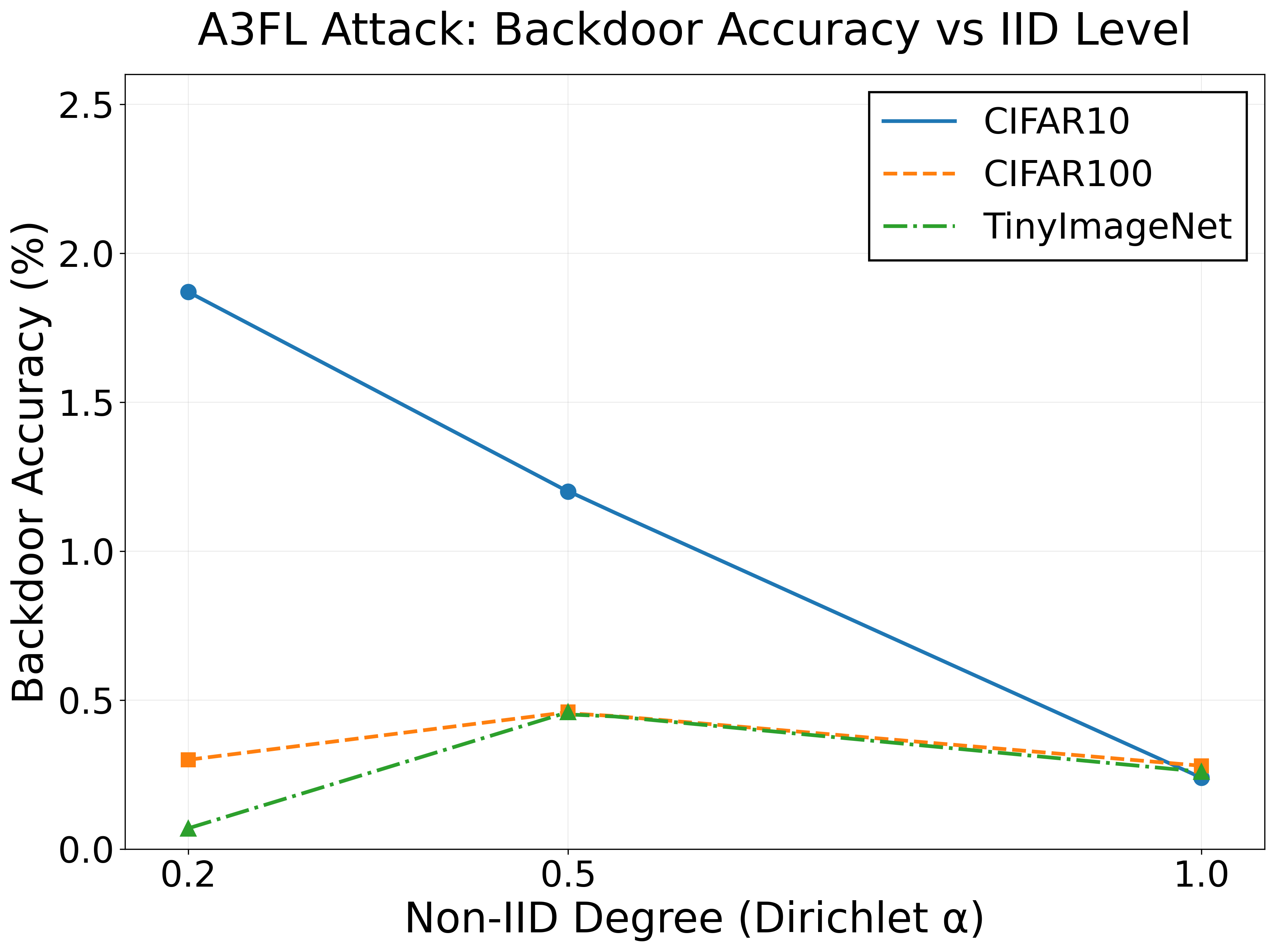}
    \caption{A3FL attack: variation of backdoor accuracy (\%) across different non-IID levels (Dirichlet $\alpha$) on CIFAR-10, CIFAR-100, and Tiny-ImageNet. FeRA consistently reduces backdoor success as data distribution becomes more uniform.}
    \label{fig:a3fl_iid_performance}
\end{figure}

We further study the effect of non-IID severity (Dirichlet parameter $\alpha$) on clean and backdoor accuracy under A3FL. Fig.~\ref{fig:a3fl_iid_performance} reports results on CIFAR-10, CIFAR-100, and Tiny-ImageNet. Across all datasets, FeRA preserves clean accuracy while strongly suppressing backdoor effectiveness as the data distribution approaches IID ($\alpha \to 1$), demonstrating robustness to client heterogeneity. On CIFAR-10, BA decreases from $1.87\%$ at $\alpha=0.2$ to $0.24\%$ under IID. On CIFAR-100, BA drops from $0.30\%$ to $0.28\%$. On Tiny-ImageNet, BA remains below $0.50\%$ for all $\alpha$, with a mild peak at $\alpha=0.5$ and a subsequent decrease under IID conditions.

\begin{table}[t]
\centering
\caption{Performance against A3FL trigger-optimization attack on CIFAR-10 under Dirichlet $\alpha = 0.5$. Results: MA(\%) / BA(\%).}
\label{tab:a3fl_cifar10}
\resizebox{0.5\columnwidth}{!}{
\begin{tabular}{lc}
\toprule
\textbf{Defense} & \textbf{MA(\%) / BA(\%)} \\
\midrule
FedAvg~\cite{mcmahan2017communication} & 86.36 / 92.47 \\
DeepSight~\cite{rieger2022deepsight} & 80.84 / 91.85 \\
RLR~\cite{ozdayi2021defending} & 75.26 / 92.36 \\
\midrule
\textbf{FeRA (Ours)} & \textbf{84.26 / 1.26} \\
\bottomrule
\end{tabular}
}
\end{table}

\subsection{Resilience Against Adaptive Attacks}

\begin{table}[t]
\centering
\caption{Defense Performance Against Adaptive BadNet Attack. The attack employs norm clipping (75\% reduction), direction alignment (30\% blend), gradual ramping (10 rounds), and adaptive scaling to evade detection metrics.}
\label{tab:adaptive_attacks}
\resizebox{0.7\columnwidth}{!}{
\begin{tabular}{lccc}
\toprule
\textbf{Dataset} & \textbf{Architecture} & \textbf{MA(\%)} & \textbf{BA(\%)} \\
\midrule
CIFAR-10 & ResNet-18 & 92.31 & 10.92 \\
CIFAR-100 & ResNet-18 & 65.88 & 0.46 \\
MNIST & MnistNet & 99.13 & 0.12 \\
EMNIST & MnistNet & 83.13 & 0.13 \\
FEMNIST & MnistNet & 78.23 & 0.46 \\
\bottomrule
\end{tabular}
}
\end{table}

We evaluate FeRA under Adaptive BadNet, a defense-aware variant that combines four evasion strategies: (1) norm clipping to suppress magnitude cues, (2) direction alignment to mimic benign trajectories, (3) gradual ramp-up over 10 rounds to avoid early detection, and (4) adaptive scaling guided by benign behavior. These mechanisms reflect realistic responses to multi-metric defenses. Table~\ref{tab:adaptive_attacks} shows that FeRA retains strong robustness across five datasets, achieving 0.12\%--10.92\% BA. On CIFAR-10, BA reaches 10.92\%, indicating that partial evasion is possible but comes at the cost of reduced backdoor strength. Simpler datasets (MNIST, EMNIST, FEMNIST) remain near-fully protected, with BA between 0.12\% and 0.46\%.

Full evasion would require simultaneously suppressing spectral and delta norms while matching benign directional and similarity patterns, which materially weakens the attack's ability to shift the global model. The aggressive clipping and alignment required to obscure detection also diminish the impact of malicious updates, leading to the observed 0.12\%--10.92\% BA. These results reinforce the value of combining complementary metrics rather than relying on a single detection signal.

\section{Limitations and Future Work}
\label{sec:limitations}

While FeRA achieves strong empirical performance across diverse attack types and datasets, our evaluation on MNIST-family datasets (Appendix~\ref{app:more_datasets}) reveals architectural dependencies. FeRA's detection efficacy degrades on shallow network architectures commonly deployed for simple datasets. Specifically, MNIST variants achieve 4.26\%--15.26\% BA compared to 1.35\%--3.68\% BA on CIFAR/Tiny-ImageNet benchmarks. This performance gap stems from limited representational depth in shallow CNNs (2--3 layers), where penultimate-layer feature dimensions ($d'$) compress both benign and malicious distributions into lower-dimensional manifolds. Such dimensional compression attenuates the spectral and spatial variance discriminability that underpins FeRA's Consistency Filter, making backdoor-induced variance suppression less distinguishable from benign low-variance patterns. Future work should investigate adaptive feature extraction strategies---such as multi-layer aggregation or dimensionality-aware thresholding---to improve detection in representation-constrained architectures.

Beyond architectural considerations, FeRA opens several promising research directions. FeRA's theoretical underpinnings in representation variance suppression require deeper analytical modeling, potentially formalizing the relationship between feature-space stability and backdoor persistence.

\section{Conclusion}
\label{sec:conclusion}

This work presented FeRA, a representation-space defense framework for federated learning that redefines the conventional detection paradigm against backdoor attacks. By focusing on behavioral consistency rather than statistical outliers, FeRA leverages inherent constraints of backdoor persistence to expose malicious clients through low-variance attention and coordinated representational patterns. The proposed dual-mechanism design, comprising consistency analysis and norm-inflation detection, captures complementary attack strategies spanning stealth-based and magnitude-based approaches that evade single-metric defenses. Extensive empirical evaluation demonstrates FeRA's robustness across datasets, attack types, and system settings, maintaining high clean accuracy and achieving significant reductions in backdoor success rates compared to state-of-the-art methods.

%\section*{Acknowledgments}

%This work was supported by EPSRC through the EnnCore project [EP/T026995/1]. Chibueze Peace Obioma was funded by Petroleum Technology Development Fund (PTDF), Abuja Nigeria, grant number PTDF/ED/OSS/PHD/CO/2079/22. We thank the reviewers for their constructive feedback.

\bibliographystyle{IEEEtran}
\bibliography{ref}

\appendices

\section{Impact of Layer Selection for Representation Extraction}
\label{app:layer_selection}

Neural networks encode information hierarchically, with early layers capturing low-level features and deeper layers encoding task-specific semantics. The choice of which layer to extract representations from will significantly impact detection performance. We conduct ablation studies comparing representation extraction from different network depths, with results presented in Table~\ref{tab:layer_selection}.

The penultimate layer (final feature representation before classifier) achieves optimal performance with 87.16\% MA and 1.35\% BA. This layer balances semantic richness, where task-relevant features concentrated in later layers effectively capture backdoor-induced representation collapse. In contrast, the mid layer achieves 90.94\% MA with 10.63\% BA---the significantly higher BA suggests mid-layer features capture less backdoor-specific variance suppression, while higher MA indicates mid-layer removals are less aggressive. Interestingly, combining multiple layers (early + mid + penultimate) yields 91.53\% MA but with 92.84\% BA, demonstrating that multi-layer concatenation creates high-dimensional spaces where malicious variance becomes harder to distinguish from benign heterogeneity. The increased dimensionality dilutes detection signals rather than enhancing them.

\begin{table}[th]
\centering
\caption{Effect of representation extraction depth in ResNet18 on defense performance against the Neurotoxin attack (CIFAR-10, Dirichlet $\alpha=0.5$). Results are reported as MA(\%) / BA(\%).}
\label{tab:layer_selection}
\resizebox{\columnwidth}{!}{
\begin{tabular}{lc}
\toprule
\textbf{Layer} & \textbf{MA(\%) / BA(\%)} \\
\midrule
Mid Layer & 90.94 / 10.63 \\
\textbf{Penultimate Layer} & \textbf{87.16 / 1.35} \\
Mid Layer + Penultimate Layer + early layer & 91.53 / 92.84 \\
\bottomrule
\end{tabular}
}
\end{table}

\section{Impact of Feature Dimension}
\label{app:feature_dimension}

Representation dimensionality affects the separability of benign and malicious patterns in feature space. We investigate FeRA's sensitivity to feature dimensions ranging from 64 to 512 for Neurotoxin attacks on CIFAR-10. As shown in Table~\ref{tab:feature_dimension}, FeRA maintains stable performance across this range, with BA varying minimally (0.63\%--1.36\%). At the lowest dimension (64), FeRA achieves 91.36\% MA with 1.36\% BA, demonstrating that even compact representations preserve sufficient discriminative information for detection. Performance peaks at 128 dimensions (0.63\% BA with 90.36\% MA), while higher dimensions (256, 512) maintain comparable results. This stability demonstrates FeRA's robustness to architectural choices regarding representation capacity.

\begin{table}[t]
\centering
\caption{Feature Dimension Impact. Neurotoxin attack on CIFAR-10, $\alpha=0.5$, poisoning from round 2001 for 100 rounds.}
\label{tab:feature_dimension}
\resizebox{0.7\columnwidth}{!}{
\begin{tabular}{lcccc}
\toprule
\textbf{$|F_{\text{Dim}}|$} & \textbf{64} & \textbf{128} & \textbf{256} & \textbf{512} \\
\midrule
MA (\%) & 91.36 & 90.36 & 91.16 & 87.16\\
BA (\%) & 1.36 & 0.63 & 1.35 & 1.35\\
\bottomrule
\end{tabular}
}
\end{table}

\section{Computational Cost Analysis}
\label{app:computational}

We evaluate FeRA's computation overhead compared with established Byzantine-robust aggregators.
FeRA computes representation-space metrics via spectral decomposition with complexity
\( O(C \times D^{2}) \), where \( C \) is the number of clients and \( D \) the representation dimension,
and computes client-parameter mutual similarity with complexity \( O(C^{2} \times P) \), where \( P \) is
the model parameter count. The detection stage adds \( O(C) \) for applying both mechanisms; benign aggregation then costs \( O(C \times P) \).
Thus, the worst-case cost is \( O(C \times D^{2} + C^{2} \times P + C) \). 
By contrast, methods such as Multi-Krum have complexity \( O(d \times N^{2}) \) and FoolsGold 
requires \( O(N^{2}) \), both scaling quadratically in the number of clients \( N \). 
Since \( D \ll P \) (for example, \( D \approx 64\!-\!512 \) and \( P \approx 10^{6}\!-\!10^{7} \)), 
the \( D^{2} \) term remains modest, while the \( C^{2} \) term applies only to similarity filtering 
and can be approximated through sampling. Empirically, on CIFAR-10 with ResNet-18 and \( N = 10 \), 
FeRA adds approximately 6.4~s per round (3.3~s for metric computation and 3.0~s for filtering), 
representing about 23\% of the total round time. All detection and filtering occur on the server side, 
introducing zero overhead on clients.

\section{Impact of Hyper-parameters}
\label{app:threshold_sensitivity}

We examine FeRA's detection performance across varying detection mechanism configurations to characterize the accuracy-security-precision trade-offs inherent in each component. For each mechanism we keep the other at its default values while varying threshold parameters to assess impact. Note that these are not hard threshold values, but rather percentile rankings.

\begin{table}[t]
\centering
\caption{Combined Score Weight Configuration Performance for CIFAR-10 under Neurotoxin Attack with Dirichlet $\alpha=0.5$.}
\label{tab:weight_ablation}
\resizebox{\columnwidth}{!}{
\begin{tabular}{ccc|cc}
\toprule
\multicolumn{3}{c|}{\textbf{Weight Configuration}} & \multicolumn{2}{c}{\textbf{Performance}} \\
\cmidrule(lr){1-3} \cmidrule(lr){4-5}
\textbf{$w_\sigma$} & \textbf{$w_\delta$} & \textbf{Description} & \textbf{MA (\%)} & \textbf{BA (\%)} \\
\midrule
0.5 & 0.5 & Equal weighting & 91.35 & 0.88 \\
0.4 & 0.6 & Spatial-prioritized & 89.99 & 0.81 \\
\textbf{0.6} & \textbf{0.4} & \textbf{Spectral-prioritized (default)} & \textbf{90.46} & \textbf{0.79} \\
\bottomrule
\end{tabular}
}
\end{table}

\begin{table*}[t]
\centering
\caption{Threshold Configuration Performance for CIFAR-10 under Neurotoxin Attack with Dirichlet $\alpha=0.5$ and 10\% Malicious Clients.}
\label{tab:consistency_ablation}
\renewcommand{\arraystretch}{1.5}
\setlength{\tabcolsep}{15pt}
\footnotesize
\begin{tabular}{ccc|ccccc}
\hline
\multicolumn{3}{c|}{\textbf{Threshold Configuration}} & \multicolumn{5}{c}{\textbf{Performance Metrics}} \\
\hline
\textbf{Combined} & \textbf{DAS} & \textbf{Mutual Sim.} & \textbf{MA (\%)} & \textbf{BA (\%)} & \textbf{Precision} & \textbf{FPR} & \textbf{TPR} \\
\hline
$\leq$50\% & $\leq$50\% & $\geq$70\% & 90.53 & 1.47 & 0.53 & 0.26 & 0.94 \\
$\leq$40\% & $\leq$40\% & $\geq$80\% & 90.63 & 3.16 & 0.74 & 0.16 & 0.94 \\
$\leq$60\% & $\leq$60\% & $\geq$60\% & 85.63 & 0.63 & 0.47 & 0.36 & 1.00 \\
$\leq$50\% & $\leq$50\% & $\geq$70\% & 90.47 & 1.36 & 0.63 & 0.16 & 0.94 \\
$\leq$50\% & $\leq$50\% & $\geq$70\% & 90.05 & 2.36 & 0.47 & 0.36 & 0.94 \\
\hline
\end{tabular}
\vspace{1mm}
\end{table*}

\textbf{Combined Score Weight Sensitivity.} The Combined Score fuses spectral and spatial attention metrics through weighted linear combination, with weights $w_\sigma$ and $w_\delta$ controlling the relative influence of each component. Table~\ref{tab:weight_ablation} evaluates three weight configurations for Neurotoxin attacks on CIFAR-10 under Dirichlet $\alpha=0.5$. Equal weighting ($w_\sigma=0.5$, $w_\delta=0.5$) achieves strong main task accuracy (91.35\%) but yields slightly elevated backdoor accuracy (0.88\%) compared to asymmetric configurations. Reversing the default weighting ($w_\sigma=0.4$, $w_\delta=0.6$) prioritizes spatial magnitude over spectral variance, achieving 0.81\% BA with 89.99\% MA. Our default configuration ($w_\sigma=0.6$, $w_\delta=0.4$) achieves optimal balance, achieving the lowest BA (0.79\%) while maintaining high MA (90.46\%). This weighting reflects the empirical observation that spectral variance suppression provides the primary discriminative signal for backdoor detection, as backdoor-induced attentional collapse manifests most strongly in eigenvalue distributions across principal components. The spatial component serves as a complementary confirmation signal, capturing attacks that attempt to evade spectral detection through directional variance concentration. The 0.6/0.4 split ensures that spectral signatures dominate detection decisions while spatial metrics provide robustness against evasion strategies targeting spectral metrics alone.

\textbf{Consistency Filter Threshold Sensitivity.} Table~\ref{tab:consistency_ablation} evaluates Combined Score, Directional Attention Score (DAS), and Mutual Similarity thresholds for Neurotoxin attacks on CIFAR-10 under Dirichlet $\alpha=0.5$ with 10\% malicious clients. Stricter thresholds (Combined $\leq$40\%, DAS $\leq$40\%, Mutual Similarity $\geq$80\%) achieve optimal precision (0.74) with minimal false positives (FPR=0.16) while maintaining 90.63\% MA, though BA increases to 3.16\% as tighter constraints allow sophisticated attacks to evade detection. Conversely, permissive thresholds (Combined $\leq$60\%, DAS $\leq$60\%, Mutual Similarity $\geq$60\%) achieve strongest attack suppression (BA=0.63\%) with TPR=1.00 but incur highest false positive rate (FPR=0.36), degrading MA to 85.63\% through excessive benign client exclusion. The default configuration (50\%/50\%/70\%) balances these objectives, achieving 1.36\%--2.36\% BA while maintaining 90.05\%--90.53\% MA and TPR near 0.94 across random seed variations.

\textbf{Norm-Inflation Filter Spectral Ratio Sensitivity.} Table~\ref{tab:fera-norm-ablation} 
examines the MAD multiplier $k$ controlling the norm-inflation detection threshold under Anticipate attacks. Conservative threshold 
($k=3$) achieves optimal precision (0.90) with minimal false positives (FPR=0.02) 
while maintaining 89.95\% MA and 0.84\% BA, effectively isolating extreme norm 
inflation without benign client interference. Aggressive threshold ($k=12$) achieves 
comparable BA (0.78\%) with perfect TPR (1.00) but reduces precision to 0.80 as 
lower thresholds flag benign variance. Moderate threshold ($k=6$, our default) 
balances these objectives, achieving 0.77\% BA with 90.32\% MA and 0.90 precision. 
The threshold range $k \in [3, 12]$ maintains BA below 0.85\% with MA above 89.95\%, 
demonstrating robust performance across Anticipate attack scenarios where norm 
manipulation dominates evasion strategy. This MAD-based approach provides automatic 
adaptation to each round's empirical distribution, eschewing fixed thresholds that 
would require manual tuning across different datasets or training dynamics.

\begin{table}[t]
\centering
\caption{Norm-Inflation Filter: MAD Sensitivity Parameter Ablation on CIFAR-10 (Anticipate, ResNet-18, Dirichlet $\alpha=0.5$, 10\% Malicious Clients). MAD Multiplier $k$ controls detection threshold.}
\label{tab:fera-norm-ablation}
\renewcommand{\arraystretch}{1.3}
\setlength{\tabcolsep}{5pt}
\begin{tabular}{lccccc}
\hline
\textbf{$k$} & \textbf{MA (\%)} & \textbf{BA (\%)} & \textbf{Precision} & \textbf{FPR} & \textbf{TPR} \\
\hline
3  & 89.95 & 0.84 & 0.90 & 0.02 & 1.00 \\
6  & 90.32 & 0.77 & 0.90 & 0.01 & 0.90 \\
9  & 90.43 & 0.80 & 0.80 & 0.03 & 1.00 \\
12 & 90.16 & 0.78 & 0.80 & 0.05 & 1.00 \\
\hline
\end{tabular}
\vspace{1mm}
\end{table}

\section{Effectiveness on More Datasets}
\label{app:more_datasets}

\textbf{MNIST-Family Evaluation.} To assess FeRA's performance across diverse visual domains, we conduct additional experiments on MNIST-family datasets including handwritten digits (MNIST), fashion products (F-MNIST), and characters (EMNIST) under DBA attacks. As shown in Table~\ref{tab:more_datasets}, FeRA achieves 4.26\%--15.26\% BA while maintaining 84.63\%--97.84\% MA, consistently outperforming FLTrust (7.16\%--23.47\% BA) and other baselines.

The elevated BA compared to CIFAR-10/100 (1.35\%--1.84\% BA) stems from shallow CNN architectures (2--3 layers) used in MNIST variants. Limited representational depth ($d' \ll 512$) compresses benign and malicious feature distributions, reducing discriminability of variance suppression signatures that drive FeRA's detection.

\begin{table}[t]
\centering
\caption{Defense performance across MNIST-family datasets (DBA attack).}
\label{tab:more_datasets}
\small
\resizebox{\columnwidth}{!}{
\begin{tabular}{l|cc|cc|cc}
\toprule
\multirow{2}{*}{Defense} & 
\multicolumn{2}{c|}{MNIST} & 
\multicolumn{2}{c|}{F-MNIST} & 
\multicolumn{2}{c}{EMNIST} \\
& BA$\downarrow$ & MA$\uparrow$ & BA$\downarrow$ & MA$\uparrow$ & BA$\downarrow$ & MA$\uparrow$ \\
\midrule
FedAvg~\cite{mcmahan2017communication} & 98.47 & 97.94 & 96.74 & 88.94 & 96.47 & 84.53 \\
FoolsGold~\cite{fung2020foolsgold} & 67.26 & 95.36 & 61.84 & 85.26 & 73.94 & 80.16 \\
DeepSight~\cite{rieger2022deepsight} & 28.63 & 96.84 & 24.36 & 87.47 & 44.16 & 82.94 \\
FLAME~\cite{nguyen2022flame} & 14.74 & 96.26 & 12.84 & 86.74 & 31.84 & 81.74 \\
Multi-Krum~\cite{blanchard2017machine} & 19.36 & 94.74 & 16.26 & 84.36 & 37.63 & 78.84 \\
FLTrust~\cite{cao2020fltrust} & 8.47 & 97.26 & 7.16 & 88.16 & 23.47 & 83.47 \\
\textbf{FeRA} & \textbf{4.26} & \textbf{97.84} & \textbf{6.94} & \textbf{88.63} & \textbf{15.26} & \textbf{84.63} \\
\bottomrule
\end{tabular}
}
\end{table}

\begin{figure*}[t]
    \centering
    \includegraphics[width=0.95\textwidth]{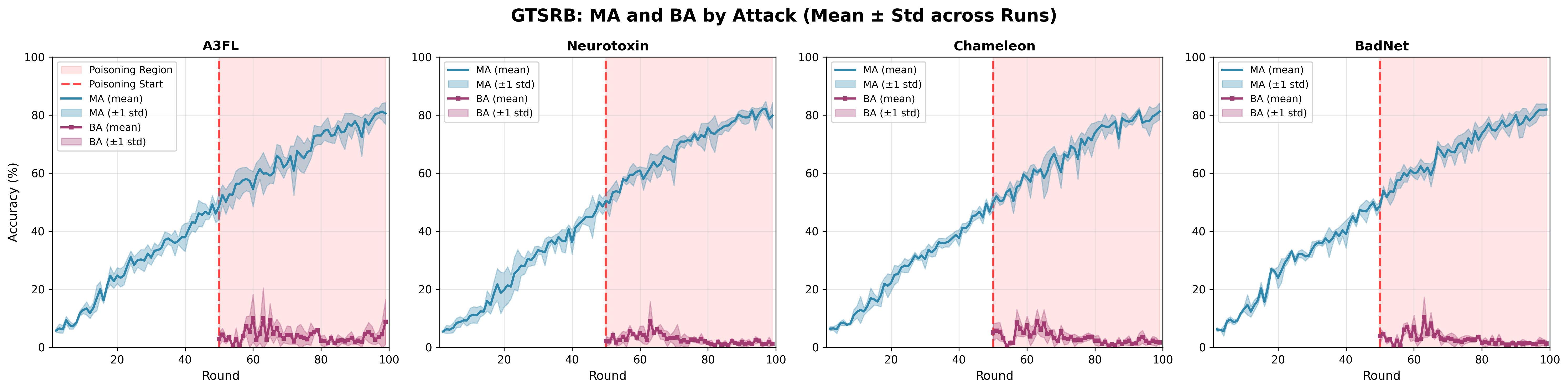}
    \caption{FeRA performanance on GTSRB dataset across four attacks over 100 training rounds. Poisoning begins at round 50 (vertical line); shaded region indicates poisoning phase. Bands show mean ± 1 std.}
    \label{fig:gtsrb}
\end{figure*}

We further evaluate FeRA on the GTSRB dataset, a real-world dataset with natural image variations and lighting conditions. Training proceeds for 100 rounds with poisoning starting at round 50. Fig.~\ref{fig:gtsrb} shows performance across four attack types.  FeRA achieves 1.20\%-4.03\% BA while maintaining 71.89\%-82.43\% MA. Against Neurotoxin, FeRA suppresses BA to 1.36\% (98.64\% reduction vs. undefended), demonstrating robust detection despite real-world feature complexity. Consistent performance across synthetic and real-world datasets validates FeRA's practical applicability for safety-critical deployments.

\section{Impact of Poisoning Data Percentage}
\label{app:poison_percentage}

\textbf{Sensitivity to Poisoning Intensity.} Attackers may vary the percentage of their local dataset that is poisoned with backdoor samples to balance stealth and effectiveness. We examine FeRA's robustness across poisoning rates from 10\% to 50\% for Neurotoxin attacks on CIFAR-10. Results in Table~\ref{tab:poison_percentage} demonstrate consistent detection across this range. For moderate poisoning (10\%--40\%), BA ranges from 1.16\% to 1.47\% with median 1.26\%, showing minimal variation despite differences in poisoning intensity. Interestingly, at 50\% poisoning, BA decreases to 0.94\% as the heavily poisoned dataset creates more pronounced representation anomalies, making detection easier. MA remains stable at 89.36\%--91.47\%, confirming utility preservation independent of poisoning rate. These findings indicate that even minimal backdoor injection (10\%) creates sufficient representation-level variance suppression for reliable identification.
\begin{table}[t]
\centering
\caption{Poisoning Percentage Impact. Neurotoxin attack on CIFAR-10, $\alpha=0.5$, poisoning from round 2001 for 100 rounds.}
\label{tab:poison_percentage}
\resizebox{\columnwidth}{!}{
\begin{tabular}{lccccc}
\toprule
\textbf{Percentage Data Poisoned} & \textbf{10\%} & \textbf{20\%} & \textbf{30\%} & \textbf{40\%} & \textbf{50\%} \\
\midrule
MA (\%) & 89.36 & 91.47 & 91.05 & 90.94 & 90.74\\
BA (\%) & 3.26 & 1.47 & 1.26 & 1.16 & 0.94\\
\bottomrule
\end{tabular}
}
\end{table}

\end{document}